\def\arxivversion{1}
\documentclass{article}
\providecommand{\arxivversion}{0}
\usepackage[utf8]{inputenc}
\usepackage[T1]{fontenc}
\usepackage[expansion=false,protrusion=true]{microtype}
\usepackage{graphicx}
\usepackage{booktabs}
\usepackage{array}
\usepackage{hyperref}
\ifnum\arxivversion=1
\usepackage[preprint]{icml2026}
\hypersetup{pdfsubject={}}
\else
\usepackage{icml2026}
\fi
\usepackage{xurl}
\usepackage{amsmath,amssymb,amsthm}
\newtheorem{proposition}{Proposition}
\usepackage{xcolor}
\usepackage[capitalize,noabbrev]{cleveref}
\usepackage{enumitem}
\definecolor{cpos}{HTML}{D9892B}\definecolor{cmark}{HTML}{D1495B}\definecolor{cseq}{HTML}{7B5EA7}
\definecolor{cread}{HTML}{2B4C7E}\definecolor{cctrl}{HTML}{9AA1AA}\definecolor{cclean}{HTML}{2A9D8F}
\newcommand{\bhfig}[3]{\IfFileExists{#1}{\includegraphics[width=#2]{#1}}{\fbox{\parbox[c][#3][c]{\dimexpr#2-2\fboxsep-2\fboxrule\relax}{\centering figure in preparation}}}}

\icmltitlerunning{Scoring Explanations Without Reading Them}

\begin{document}

\twocolumn[
  \icmltitle{Scoring Explanations Without Reading Them: Structural Shortcuts in\\
    Automated-Interpretability Evaluation Data}

  \ifnum\arxivversion=1
\begin{icmlauthorlist}
  \icmlauthor{Xupeng Chen}{dg}
\end{icmlauthorlist}
\icmlaffiliation{dg}{Dimension Gate}
\icmlcorrespondingauthor{Xupeng Chen}{xc1490@nyu.edu}
  \else
  \begin{icmlauthorlist}
    \icmlauthor{Anonymous Authors}{anon}
  \end{icmlauthorlist}
  \icmlaffiliation{anon}{Anonymous Institution}
  \icmlcorrespondingauthor{Anonymous Authors}{anonymous@example.com}
  \fi
  \icmlkeywords{automated interpretability, sparse autoencoders, evaluation, shortcuts}

  \vskip 0.3in
]

\printAffiliationsAndNotice{}

\begin{abstract}
Can an automated-interpretability score be earned without reading the explanation it judges? Earlier checks varied explainers, models or settings, not item construction. We score four public
pipelines with rules reading only this construction. On Neuronpedia's stored fuzzing runs, a seed-free rule
that counts marked tokens, told the negatives' mark count (which the constructor computes from the positives'
activations), reaches balanced accuracy 0.969, against 0.838 for stored LLM scorers; reading only a run's unlabeled
marks, a post hoc rule reaches 0.905 on GPT-2 and, frozen, reaches 0.902 against 0.844 on 500 held-out features in a preregistered test;
it is not resolved on 27 Gemma runs. The pinned scorer fork, run on all 2{,}030 stored runs, gives our rebuilt marks on every item
and reproduces each run's stored item order; on these marks a centre-token rule reaches 0.979. Current delphi and
OpenAI's simulation sets carry similar cues. Jev, a typed decision model, scores higher on stored snippets than on
natural windows for 44 of 49 features; a preregistered test does not show the cue reordering features. On natural
windows cut at random offsets, Jev's mean pooled correlation is 0.331 with the real explanation and 0.013 with
shuffled ones; an open 8B simulator also reads it. Removing the mark cues moves neither the stored llama scorer's level
nor the one published dictionary gap we could rescore. Score levels here say little about the explanation,
though published orderings are not shown to be wrong. Papers that compare dictionaries or explainers should report each score beside the best explanation-free rule on the same items.
\end{abstract}

\section{Introduction}\label{sec:intro}

\begin{figure}[!t]
\centering
\includegraphics[width=\columnwidth]{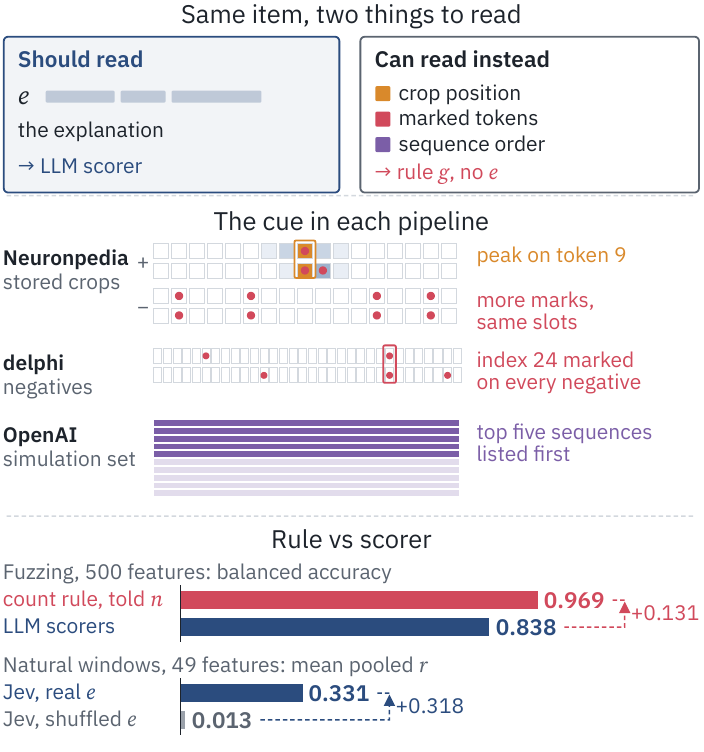}
\caption{\textbf{A rule that counts marks beats the stored LLM scorers; without the cues, Jev scores with the real
explanation and not with a shuffled one.} \emph{Top:} a scorer should read the explanation $e$ (words hidden); a rule
$g$ can read how the item was built. \emph{Middle:} the cue in each pipeline as token boxes; strips show the measured
structure, not one stored item. \emph{Bottom:} fuzzing balanced accuracy over
500 GPT-2 small features, count rule 0.969 (told $n$, the negatives' mark count) against stored LLM scorers 0.838 ($+0.131$). On natural
windows over 49 features, Jev's mean pooled $r$ is 0.331 with the real explanation and 0.013 with a shuffled one ($+0.318$). Values:
\cref{sec:cues,sec:clean}.}
\label{fig:arc}
\end{figure}

A researcher who trains a new sparse-autoencoder (SAE) dictionary often compares it with older ones by how well an
LLM scorer judges its features' explanations. Open a GPT-2 small \citep{radford2019gpt2} feature on Neuronpedia
\citep{neuronpedia} and look at a stored fuzzing run. Its twenty activating examples are 17-token windows cut around the feature's strongest activation, so
the peak sits on the ninth token. Its five non-activating examples carry marks placed by a seeded sampler. The scorer,
an LLM, reads the explanation and each example with some tokens marked. It answers whether the marked tokens are what
the explanation describes \citep{paulo2024autointerp}.

These scores are what SAE papers and benchmarks use to compare dictionaries and explainers
\citep{paulo2024autointerp,karvonen2025saebench}. Their items are built by code that decides where a window is
cropped, which tokens are marked and how many, and which sequences come from the top of the activation distribution. Each choice can tell a rule the answer without the explanation, and a score that such a rule matches is not, by
itself, evidence about the explanation.

Three studies are closest. \citet{vanderben2026building} vary pipeline settings and find that they move scores more
than SAE architecture does, with fuzzing the least stable metric. A cue present in every setting adds no variance, so that design cannot see it. \citet{heap2025random} and
\citet{korznikov2026sanity} find that SAEs on random transformers, or with random feature directions, score close to
trained ones: the metrics are insensitive to the model. \citet{paulo2025noexplanations} score latents without
explanations to avoid the explainer. These studies vary the explainer, the model or the settings; none measures what
the construction of the items alone lets a rule score.

\begin{figure*}[t]
\centering
\includegraphics[width=\textwidth]{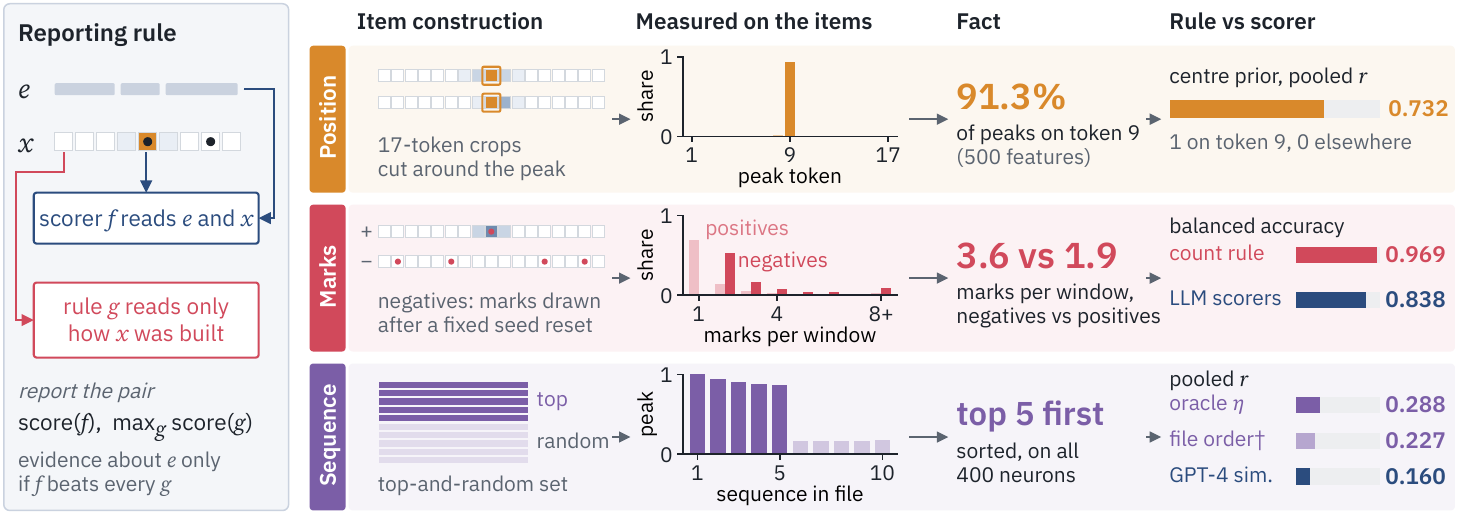}
\caption{\textbf{Each cue is one fact about how items are built, and a rule that reads only that fact is scored
beside the scorer.} \emph{Left:} the reporting rule of \cref{eq:gap,eq:evidence}: scorer $f$ reads $e$ and $x$, rule $g$
only how $x$ was built. \emph{Right:} one row per cue. \emph{Position:} 91.3\% of activating
examples peak on token 9 (500 features), and the centre prior reaches pooled $r$ 0.732. \emph{Marks:} over 1{,}001 stored
runs, negatives carry 3.6 rebuilt marks and positives 1.9; the count rule reaches balanced accuracy 0.969 against 0.838
for the LLM scorers. \emph{Sequence:} OpenAI's files list the top five sequences first (400 neurons); the sequence-mean
oracle $\eta$ reaches 0.288 and the file-order rule$^\dagger$ 0.227, against 0.160 for GPT-4's simulator. Histogram: each
feature's first stored run; sequence panel: peak activation over the neuron's maximum. Values: \cref{sec:cues}.}
\label{fig:protocol}
\end{figure*}

We therefore score the items of four public pipelines, and a control we build, with rules that read only how an item was made (\cref{fig:arc,fig:protocol}). Such a rule sees token positions, the mark set and which sequence a token belongs to. Three of
the four pipelines carry a cue such a rule can read, and each cue has a closed form or a sharp bound
(\cref{sec:setup}). Analyses chosen after seeing the data carry a dagger ($^\dagger$).

A rule that never reads the explanation and only counts the marked tokens
is right on almost every item when told the negatives' mark count, which the constructor computes from the
positives' activations. For 500 random features and 1{,}001 stored runs,
that rule, which does not use the seed, reaches balanced accuracy 0.969, against 0.838 for the stored LLM scorers (recomputed from their
stored per-item verdicts); from a run's unlabeled marks alone, 0.905$^\dagger$, and 0.902 on 500 held-out features (preregistered), unresolved on Gemma.
The runs do not store which tokens were marked, so we rebuild them; the pinned scorer fork, run on all 2{,}030 stored runs, gives the same
marks on every item and reproduces their stored item order (\cref{app:provenance}). On them, a rule
answering ``yes'' when the ninth token is marked reaches 0.979, falling to 0.880 with the negatives' positions drawn at random.

Whether the cue changes conclusions is a separate question, and our answer concerns levels, not orderings. We test it
with one scorer, Jev, a commercial typed decision model that returns a distribution over declared answer options
instead of generating text \citep{typesafe2026jev}. Its simulation scores are higher on stored snippets than on
natural windows for 44 of 49 features$^\dagger$, sets that also differ in window length and composition. In a
crop-by-length factorial the gap follows length, not centring. A small scorer trained on the stored snippets with no
explanation at all reaches 83\% of the score of the same model trained with one. A preregistered test finds Jev's
scores tracking crop geometry but does not show the cue reordering features (Spearman 0.732 between the two sets;
\cref{sec:harm}). So a score's level on these data says little about the explanation, and published orderings of features
are not shown to be wrong.

This changes reporting: a score is evidence about an explanation only if the scorer beats every
explanation-free rule on the same items (\cref{eq:gap}; four checks in
\cref{sec:conclusion}).

\paragraph{Contributions.}
\begin{enumerate}[leftmargin=*,topsep=2pt,itemsep=1pt]
\item \textbf{Diagnosis.} Fuzzing negatives give away their marks, stored crops the peak's position, and
top-and-random sets the sequence order. A seed-free count rule told the negatives' mark count reaches 0.969 against the stored scorers' 0.838;
closed forms bound what such rules score (\cref{sec:setup,sec:cues}).
\item \textbf{Consequence.} One scorer's simulation scores are higher on stored snippets than on natural windows, in
sets that differ in length and composition; the gap follows length, $+0.084$, not centring, $-0.006$. A preregistered
test does not show the cue reordering features. On 1{,}000 stored runs one of two LLM scorers accepts centre-unmarked
positives less often (item not held fixed), and neither follows the count cue (\cref{sec:harm}).
\item \textbf{Remedy.} Random-offset windows with the beginning-of-sequence (BOS) token remove the position cue; matched marks remove the mark
cues without moving the stored llama scorer's level; $r_w$ separates the sequence cue. On natural windows Jev beats every
position prior and its shuffled control; its margin over the sequence-mean oracle does not survive Holm correction.
\end{enumerate}

\section{Related Work}\label{sec:related}

\paragraph{Explaining and scoring units automatically.}
Unit description began in vision \citep{bau2017dissection,hernandez2022milan,oikarinen2022clipdissect}.
\citet{bills2023neurons} introduced explain-then-simulate for neurons, and SAE work adopted it
\citep{cunningham2023sae,bricken2023monosemanticity,templeton2024scaling}. \citet{paulo2024autointerp} scaled it and
added detection and fuzzing, released as delphi \citep{eleuther2024delphi,eleuther2024autointerp}, whose fork
Neuronpedia runs \citep{neuronpedia}. Other work tunes explainers and simulators \citep{choi2024scaling}, improves
explanations \citep{singh2023sasc,gurarieh2025output} or uses agents \citep{shaham2024maia}. We fix both.

\paragraph{Evaluating explanation scores and SAEs.}
A high simulation score can come with an explanation that misses a unit's causal role
\citep{huang2023rigorous}. Explanations were checked on functions with known descriptions
\citep{schwettmann2023find}, generated concept data \citep{kopf2024cosy} and sanity checks
\citep{oikarinen2025unified}; FADE separates failure modes \citep{puri2025fade}, and explanations depend on
the data used to find top activations \citep{bolukbasi2021illusion}. Benchmarks add task-based SAE evaluation
\citep{makelov2024principled,karvonen2025saebench,kantamneni2025probing}. The closest studies are
compared in \cref{sec:intro}; pipeline settings also change top-$k$ feature rankings \citep{vanderben2026building}. These vary the explanation, model, task or settings; we hold them fixed and measure what item construction alone lets a rule score. Our ranking test (\cref{sec:harm}) is one
scorer's rank correlation between two sets, not top-$k$ stability across draws.

\paragraph{Shortcuts and leakage in evaluation data.}
Benchmarks often admit solutions that ignore the intended signal: hypothesis-only baselines and
annotation artifacts \citep{poliak2018hypothesis,gururangan2018annotation}, syntactic heuristics
\citep{mccoy2019right}, shortcut learning \citep{geirhos2020shortcut}, saliency maps that ignore the
model \citep{adebayo2018sanity}, and leakage \citep{kapoor2023leakage}. Our rules are
input-only baselines that read the item's construction, not its text, so they have closed forms.

\paragraph{Typed decision models and cheap judges.}
LLMs serve as annotators and judges with position biases
\citep{gilardi2023chatgpt,ziems2024css,zheng2023judging,wang2023fair}; cost-aware systems route easy queries to cheaper
models \citep{chen2023frugalgpt,ding2024hybrid,ong2024routellm}, and encoders score predefined options cloze-style
\citep{schick2021pet,devlin2019bert,warner2024modernbert,marone2025mmbert}. Typed decision models are
cheaper and less accurate than frontier LLMs for annotation \citep{ibrahim2026decision}, have an open implementation
\citep{laya2026}, and can follow an option's name, not its definition \citep{sun2026typesafe}. Calibration depends on
model and format \citep{guo2017calibration,zhao2021calibrate,kadavath2022know}. We use one only as a scorer that meets \eqref{eq:evidence}, not as a recommended one.

\section{Problem Setup}\label{sec:setup}

\paragraph{Simulation.} A feature is scored on sequences $s=1,\dots,m$ with tokens $t=1,\dots,n_s$ and
true activations $a_{st}$; $N=\sum_s n_s$. A simulator reads the explanation $e$ and returns $\hat
a_{st}$ for every token. The score is the Pearson correlation over all tokens pooled, as in OpenAI's
released code \citep{bills2023neurons}:
\begin{equation}\label{eq:score}
r(a,\hat a)=\frac{\sum_{s,t}(a_{st}-\bar a)(\hat a_{st}-\bar{\hat a})}
{\big[\sum_{s,t}(a_{st}-\bar a)^2\big]^{1/2}\big[\sum_{s,t}(\hat a_{st}-\bar{\hat a})^2\big]^{1/2}}.
\end{equation}
\paragraph{Fuzzing and detection.} An item $x$ is a window, positive if the feature is active on it and
negative otherwise, with a set of marked token positions $M(x)$. Fuzzing asks whether the marked tokens
are instances of $e$; detection shows the window unmarked and asks whether any token is
\citep{paulo2024autointerp}. A verdict $g(x)\in\{0,1\}$ is scored by balanced accuracy
\begin{equation}\label{eq:ba}
\mathrm{BA}(g)=\tfrac12\big(\Pr_+[g=1]+\Pr_-[g=0]\big),
\end{equation}
with $\Pr_\pm$ over the positive and negative items.

\paragraph{Explanation-free rules and the estimand.} A rule is explanation-free if it is a function of
the item's construction alone: token positions, the mark set $M(x)$, and which sequence a token belongs
to, but not $e$ and not the text. For a scorer $f$ and a pre-specified family $\mathcal G$ of such
rules evaluated on the same items, we report the pair
\begin{equation}\label{eq:gap}
\big(\mathrm{score}(f),\ \max_{g\in\mathcal G}\mathrm{score}(g)\big).
\end{equation}
We call a score evidence about $e$ only when
\begin{equation}\label{eq:evidence}
\mathrm{score}(f;e)>S_{\mathcal G}\ \ \text{and}\ \ \mathrm{score}(f;\tilde e)\le S_{\mathcal G},
\end{equation}
with $S_{\mathcal G}=\max_{g\in\mathcal G}\mathrm{score}(g)$ and $\tilde e$ a shuffled explanation. The pair states what a score certifies about $e$, not that $f$ used a cue;
whether a scorer follows a cue is measured separately, holding the run and the label fixed (\cref{sec:harm}). The
families we use are the fixed-position priors (1 at one token index, 0 elsewhere), the centre-marked and fixed-slot mark
rules, and the count rule ``yes iff $|M(x)|$ differs from the negatives' mark count'', a run parameter the rule is told. Beside them we report, as a
ceiling rather than a rule, the sequence-mean oracle $\hat a_{st}=\bar a_s$. It uses the true sequence means and bounds
every rule constant within sequences (\cref{prop:affine}(c)). Post hoc rules ($^\dagger$) are optimistic.

\paragraph{Three closed forms.} Known results \citep{robinson1950ecological,tate1954pointbiserial,lehmann2005testing}
(\cref{app:proofs}).

\begin{proposition}[Sequence-level predictors]\label{prop:affine}
Assume $a$ and $\hat a$ are not constant. Let $\bar a_s,\bar{\hat a}_s$ be per-sequence means, let
$\eta=\big(\sum_s n_s(\bar a_s-\bar a)^2/\sum_{s,t}(a_{st}-\bar a)^2\big)^{1/2}$ be the correlation
ratio of $a$ and $\hat\eta$ that of $\hat a$, let $r_w$ be the correlation of the within-sequence
residuals $a_{st}-\bar a_s$ and $\hat a_{st}-\bar{\hat a}_s$ pooled over all tokens, and let $r_b$ be
the $n_s$-weighted correlation of the sequence means $\bar a_s$ and $\bar{\hat a}_s$ (in (b), the
first product is read as zero if $\|w\|\|\hat w\|=0$, and the second if $\eta\hat\eta=0$).
(a) For every $\alpha>0$ and $\beta\in\mathbb R$, $r(a,\alpha\hat a+\beta)=r(a,\hat a)$; for increasing
maps that are not affine on the values of $\hat a$ this fails in general.
(b) $r(a,\hat a)=r_w\sqrt{1-\eta^2}\sqrt{1-\hat\eta^2}+r_b\,\eta\,\hat\eta$; in particular the
between-sequence contribution $r_b\eta\hat\eta$ is at most $\eta$.
(c) Every predictor that is constant within sequences scores $r_b\eta\le\eta$. If $\eta>0$, equality
holds if and only if its sequence values are a positive affine function of $\bar a_s$, and the
sequence-mean predictor $\hat a_{st}=\bar a_s$, which is given each sequence's true mean activation and
nothing about tokens, scores exactly $\eta$. For any $\hat a$ and any non-constant sequence-level offset $\delta_s$,
$r(a,\hat a+\lambda\delta_{s(\cdot)})\to r(a,\delta_{s(\cdot)})\le\eta$ as $\lambda\to\infty$.
\end{proposition}

\begin{proposition}[The centre prior]\label{prop:centre}
Assume some $n_s\ge2$. Let $c_{st}=1$ at the centre token $t=\lfloor n_s/2\rfloor+1$ of each
sequence (0-based index $\lfloor n_s/2\rfloor$) and $0$ elsewhere, let $\pi=m/N$ be the fraction of
centre tokens (so $0<\pi<1$), $\mu_c,\mu_o$ the mean of $a$ over centre and other tokens, and
$\sigma_a>0$ the pooled population standard deviation of $a$ (weight $1/N$). Then
$r(a,c)=(\mu_c-\mu_o)\sqrt{\pi(1-\pi)}/\sigma_a$.
If moreover every non-centre token has $a_{st}=0$, $\mu_c>0$, and $\kappa=s_c/\mu_c$ with $s_c^2$ the
population variance (weight $1/m$) of the $m$ centre activations, then
$r(a,c)=\sqrt{(1-\pi)/(1-\pi+\kappa^2)}$.
\end{proposition}

\begin{proposition}[Rules that read only the marks]\label{prop:marks}
Let $S(x)$ be any statistic of an item's construction (for instance the mark set $M(x)$, its size, or
the indicator that a given position is marked), with laws $P_+$ and $P_-$ on positives and negatives,
densities $p_\pm$ with respect to a common dominating measure $\nu$, and
$\mathrm{TV}(P_+,P_-)=\sup_A|P_+(A)-P_-(A)|=\tfrac12\int|p_+-p_-|\,d\nu$. Let $g$ be any rule that
depends on $x$ only through $S$, possibly randomised with randomness independent of the item (then
$\Pr_\pm$ in \eqref{eq:ba} also average over that randomness). Then
$\mathrm{BA}(g)\le\tfrac12\big(1+\mathrm{TV}(P_+,P_-)\big)$, with equality for
$g^\star(S)=\mathbf 1[p_+(S)>p_-(S)]$. If $P_+=P_-$, every such rule has $\mathrm{BA}(g)=\tfrac12$ exactly.
\end{proposition}

By \cref{prop:affine}(c) a simulator that gets only each sequence's level right reaches $\eta$ on a
top-and-random set. Part (a) is why a typed simulator's expected level needs no linear calibration, and why OpenAI's
calibration step leaves the score unchanged whenever the fitted slope is positive \citep{bills2023neurons}. \Cref{prop:centre} fixes the centre
prior's score by $\mu_c-\mu_o$, $\pi$ and $\sigma_a$, none involving the explanation. \Cref{prop:marks} makes
matched mark laws a design rule (\cref{sec:clean}): a scorer then beats chance only through what it reads
beyond the marks.

\paragraph{Status and statistics.} The 500-feature, fuzzing-rule, delphi and SAEBench analyses and the harm test
were preregistered with decision rules (\cref{app:prereg}), as were the cue-use, cross-dictionary,
explanation-choice, explainer, trained-scorer, matched-mark, natural-window, open-simulator, crop-by-length and GPT-4
comparisons. The centre prior and
$\eta$ on the 50 stored-snippet features and the 400 OpenAI neurons predate them, and post hoc analyses are marked
$^\dagger$. Unless stated otherwise, $p$-values are two-sided sign tests over runs, features or
neurons, not adjusted for multiple comparisons. Holm correction over the 19 in the main text keeps all below 0.05
except three (0.0094, 0.014, 0.044).

\section{Can the Items Be Scored Without the Explanation?}\label{sec:cues}

\textbf{We measure four public pipelines and one control.} (i) Neuronpedia's stored fuzzing runs for the
residual-stream SAE on GPT-2 small layer 8 \citep{bloom2024gpt2saes}, the only SAE with stored runs at scale that we
found. We use 500 features
drawn at random (the 500-feature sample), with 1{,}001 runs by LLM scorers \citep{geminiteam2024gemini15,grattafiori2024llama3}.
We also use the first 50 features in index order, which we scored ourselves. From Gemma-2-2B
\citep{gemmateam2024gemma2} we take every layer-20 Gemma Scope feature \citep{lieberum2024gemmascope} among indices 0--1{,}199
that carries a stored run: 12 features, 27 newer runs (run dates in \cref{app:provenance}). (ii) Examples built
locally for the 50 features with our re-implementation of the example constructor and fuzzing sampler of delphi (EleutherAI/delphi at
commit \texttt{4fea06e}), in its default mode and with centring enabled. (iii) SAEBench's scoring windows, built
the same way. (iv) The 400 GPT-2 XL neurons we sampled from OpenAI's release, with GPT-4's \citep{openai2023gpt4}
explanations, top-and-random scoring sequences, true activations and per-token simulations \citep{bills2023neurons}. (v) As a control, natural windows
we built (\cref{sec:clean}).

\textbf{Stored runs keep the verdicts but not the marks, so we rebuild the marks.}  Stored accuracies below are recomputed from those verdicts (the score Neuronpedia
displays differs on some runs, \cref{app:provenance}). Neuronpedia's server still runs a fork of delphi's predecessor,
\texttt{sae-auto-interp}. We rebuild the marks with our re-implementation of that
fork's rule. Positives mark every token at or above 0.3 of the item's maximum, and negatives a sample of
fixed size drawn after resetting the seed to 22. The fork pinned at commit \texttt{05d5718} gives these marks and prompts exactly
and, through its seeded shuffle, each run's stored item order (\cref{app:provenance}). Positives' marks and the
negatives' mark count follow from the stored activations, and the count rule does not use positions. With the
negatives' positions drawn uniformly at random instead (exact expectation)$^\dagger$, the centre rule scores 0.880
against the stored scorers' 0.838, higher on 523 of 1{,}001 runs ($p=0.16$).

\subsection{Fuzzing negatives give away their marks}\label{sec:cues:marks}

\textbf{Counting the marks beats the stored scorers.} The fork marks each negative with $n$ tokens, where $n$ is the
mean number of active tokens on the positives rounded up. Active tokens outnumber the tokens at or above 0.3 of the
maximum that positives mark. The positions are drawn after resetting the seed, so negatives are marked at the same
slots. They also carry more marks than positives: 3.6 against 1.9 on the GPT-2 runs, 19.9 against
9.2 on the Gemma runs. On the GPT-2 runs the count rule reaches 0.969 (\cref{fig:results}a). On the Gemma runs
the count rule reaches 0.981 on rebuilt marks against 0.794 for the stored scorers (the rule wins on 24 of 27 runs,
loses on 2, $p=1.0\times10^{-5}$).

\textbf{Without $n$, mark-only rules beat the stored scorers on held-out GPT-2 features, not on Gemma.} Reading only a run's
unlabeled marks, ``yes'' below the run's largest mark count reaches 0.905 on GPT-2 and 0.837 on 27 Gemma runs ($+0.044$ [$-0.134$, 0.171]); from one item, ``yes'' if its marks form one span gives 0.924 and 0.730, below Gemma's stored 0.794 (\cref{app:provenance}). Both were found post hoc$^\dagger$. Frozen, the first
reaches 0.902 against 0.844 on 500 held-out features, the second 0.913 (preregistered; intervals in \cref{app:provenance}).

\textbf{Current delphi marks every negative at the same index.} Current delphi removes the crop by default:
the flag \texttt{center\_examples=True} runs the un-centred windows (\cref{app:provenance}), so only 2.8\% of test peaks fall at index 24 of 32. The
fixed-index simulation prior then scores $-0.003$. But delphi's sampler always highlights index
$\ell-\lfloor\ell/4\rfloor=24$ in a negative, plus seeded companion marks (every negative here gets at
least two). Every negative is marked at 24, against 4.3\% of positives, so ``yes iff
index 24 is not marked'' reaches 0.978 and the count rule 0.944 on these items. Centring the windows (flag
\texttt{False}) cancels the index-24 rule, but a companion-mark rule chosen after the fact still gives
0.994$^\dagger$ (\cref{app:provenance}). SAEBench scores with centred 21-token windows (180 of 200 peaks at index 10)
but shows them unmarked for detection \citep{karvonen2025saebench}, so no mark rule has an input.

\subsection{Cropped windows give away the peak's position}

\textbf{Most stored windows peak on the centre token.} Neuronpedia's stored examples are 17-token windows centred
on the activation peak: over the 500 random features, 91.3\% of activating examples peak on the centre
token. A simulator that predicts 1 on the centre token and 0 elsewhere reaches a mean pooled
correlation of 0.732, and the best fixed index, chosen after the fact, 0.760.

\textbf{On rebuilt marks, a centre rule also beats both stored scorers.} The rebuilt positive marks almost always
include the centre and the fixed-seed negative marks almost never do. So the rule ``yes iff the centre token is
marked'' reaches balanced accuracy 0.979 on rebuilt marks. That is above
both stored scorers on the same runs: gemini-2.0-flash 0.864 (the rule wins on 312 runs and loses on 59,
$p=1.1\times10^{-42}$) and llama-3.1-70b 0.812 (437 to 23, $p=2.7\times10^{-100}$). Unlike the count rule, this one depends on the seed, which the stored item order confirms. The first 50 features agree (\cref{app:det}).

\textbf{Full-context runs carry no position cue.} The Gemma runs use 127-token windows peaking anywhere: the centre prior is 0.003 and the centre rule 0.552 on rebuilt marks.

\subsection{Top-and-random sets give away sequence order}

\textbf{Knowing each sequence's mean activation beats GPT-4's simulator.} OpenAI's scoring sets mix top-activating
sequences with random ones \citep{bills2023neurons}. By \cref{prop:affine}(c), the sequence-mean
predictor, given each sequence's true mean activation, scores the
correlation ratio $\eta$, the ceiling of every rule constant within sequences. On our 400 neurons
$\eta$ is 0.288 on average, above GPT-4's published simulator (0.160).

\textbf{The file order alone recovers most of it.} The release lists each neuron's five top-activating
sequences first (sorted on all
400 neurons, above every later sequence on 398). Predicting 1 on those five scores 0.227$^\dagger$, above GPT-4
on 287 of 400 neurons ($p=1.5\times10^{-18}$). GPT-4's simulator did not see this order; the
rule shows what the data carry.

\textbf{A pooled score hides how much credit comes from between sequences.} Such credit is not a defect (an
explanation naming the top sequences' contexts earns it), but a pooled score hides that share.
\Cref{prop:affine}(b) splits off $r_w$, which no sequence-level description reaches.

\begin{figure*}[t]
\centering
\includegraphics[width=\textwidth]{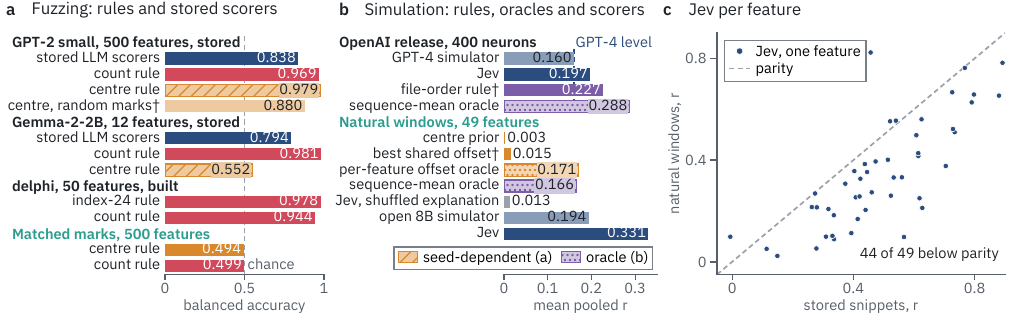}
\caption{\textbf{The best explanation-free rule beats the stored scorers on every cued set; on clean windows the shared
position priors fall to near zero and Jev with the real explanation scores highest.} (a)~Fuzzing balanced accuracy on
the stored GPT-2 and Gemma runs, delphi's built items and our matched-mark items (count rule told $n$); hatched: the centre rule, which depends on the fork's seed; dashed: chance. (b)~Simulation, mean pooled $r$, on OpenAI's 400 neurons and our 49
natural-window features; dotted: oracles given true activations; dashed: GPT-4's simulator. (c)~Jev per feature on
stored snippets against natural windows; 44 of 49 lie below parity. $^\dagger$Chosen after seeing the data. Values:
\cref{sec:cues,sec:harm,sec:clean,sec:h2h}.}
\label{fig:results}
\end{figure*}

\section{Do the Cues Change What the Scores Say?}\label{sec:harm}

\textbf{The preregistered test does not establish ranking harm; its sensitivity on real data is unmeasured.} We tested
whether the cue changes how scores rank features, on the 49 features that
have both stored snippets and natural windows, with Jev's simulation scores on both (Neuronpedia stores
no simulation scores). The verdict rule was fixed in advance:
``harm'' if the Spearman correlation of stored scores with the centre prior ($\rho_{\mathrm{geo}}$)
exceeds their correlation with natural-window scores ($\rho_{\mathrm{nat}}$) with a bootstrap interval
excluding zero. Jev's stored-snippet scores do track crop geometry,
$\rho_{\mathrm{geo}}=0.512$ [0.23, 0.72], but they track its natural-window scores more closely,
$\rho_{\mathrm{nat}}=0.732$ [0.53, 0.87], below the split-half ceilings of the two sets (0.925 and
0.888). The difference is $-0.221$ [$-0.52$, 0.06], and the partial correlation of stored with natural
scores given the centre prior is 0.710 [0.52, 0.85]. The preregistered verdict is ``geometry confound
present, ranking harm not shown'' (the same rule fires on synthetic rankings built mostly from crop
geometry, \cref{app:protocol}). When a centre cue was planted on real data, the rule did not fire (\cref{sec:limits}).

\textbf{Scores are higher on the cued snippets, and the gap follows window length.} Jev's mean score falls from
0.478 on stored snippets to 0.331 on natural windows, lower for 44 of 49 features
($p=7.6\times10^{-9}$; \cref{fig:results}c)$^\dagger$. The two sets also differ in window length (17 against 64
tokens) and composition (5 of 25 stored snippets are non-activating; 5 of 10 natural windows are random).
This comparison therefore does not isolate the cue. A crop-by-length factorial (Jev, 43 features; \cref{app:protocol}) puts
the gap on length (17 over 64 tokens: $+0.084$; centring: $-0.006$).

\textbf{One of two stored scorers accepts centre-unmarked positives less often.} On the 500-feature runs
(first-50 correlations in \cref{app:dicts}), gemini-2.0-flash accepts the 295 positives whose centre is unmarked 24 points less often than centre-marked
positives of the same run (Mantel--Haenszel difference $-0.244$ [$-0.33$, $-0.16$], $p=1.4\times10^{-21}$). For
llama-3.1-70b the difference is $-0.036$ [$-0.10$, 0.03] ($p=0.28$). Neither accepts the 613 positives carrying the
negatives' mark count less often ($+0.050$, $-0.005$), and both side with the label on most disagreeing items
(0.64, 0.70). The comparison fixes the run and the label, not the item. Off-centre peaks may be harder to judge,
but on the same items llama-3.1-70b dips by only 0.036, so item differences would have to affect gemini-2.0-flash
far more$^\dagger$.

\textbf{One explainer comparison keeps its sign; the cue level runs against one dictionary gap.}
The explainer comparison the set allows
keeps its sign. Explanations by gpt-4o-mini beat those by gpt-3.5-turbo by 0.098 on stored snippets (32 of 49,
$p=0.044$) and by 0.062 on natural windows (30 of 49, $p=0.15$; 10-level rubric: 0.079, 0.058). The gap is larger on the
cued data. We also rebuilt items from Neuronpedia's
hosted examples of five SAEs. On the one pair with a published fuzzing gap (Gemma Scope 9B, 131k
minus 16k), the count rule's level differs by $-0.028$ [$-0.041$, $-0.015$] against a published $+0.03$
\citep[Table A10]{paulo2024autointerp}. The two are close in size and opposite in sign (``not resolved'', \cref{app:dicts}), so a
scorer following this cue would narrow that ordering, not produce it. \Cref{sec:clean} rescores this pair with the
mark cues removed.

\textbf{Explanation choice is unresolved; a trained scorer misses the shortcut criterion.} Whether Jev's best explanation on stored snippets is also best on natural windows cannot be resolved at ten windows per feature (a re-ordering is neither shown nor excluded); a small encoder trained on the stored snippets misses the preregistered shortcut criterion, half its score kept with a blank explanation (\cref{app:further}).

\begin{figure*}[t]
\centering
\includegraphics[width=\textwidth]{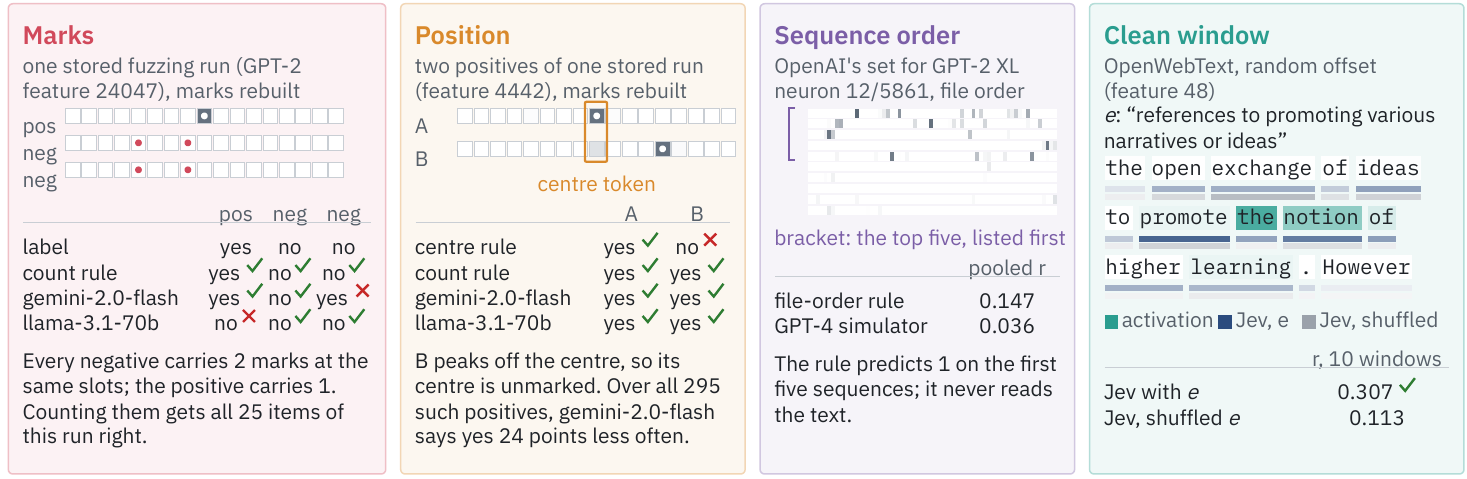}
\caption{\textbf{On real items, counting marks gets a whole stored run right, the centre rule fails an off-centre peak,
file order outscores GPT-4, and on a clean window Jev reads the explanation.} Stored items: token boxes shaded by activation (no snippet text), rebuilt marks as dots. \emph{Marks:} in one stored run (GPT-2 feature 24047) every
negative carries 2 marks at the same slots and every positive 1, so the count rule gets all 25 items right. \emph{Position:} a positive of feature 4442 peaks off the
centre. \emph{Sequence order:} OpenAI's set for GPT-2 XL neuron 12/5861, in file order; the file-order
rule reaches $r=0.147$, GPT-4's simulator 0.036. \emph{Clean window:} an OpenWebText window cut at a random offset for
feature 48, with its true activation and Jev's level under the real and a shuffled explanation. Over the feature's 10
windows, Jev's $r$ is 0.307 and 0.113. Case rules: \cref{app:provenance}.}
\label{fig:cases}
\end{figure*}

\section{Can the Cues Be Removed or Separated?}\label{sec:clean}

\textbf{Matched marks remove the fuzzing cues without moving the stored scorer's level.} By \cref{prop:marks}, if
each negative's mark set is drawn from the positives' law independently of its label and text, every mark-only rule has
expected balanced accuracy $\tfrac12$. It is exactly $\tfrac12$ if the positives' mark sets are reassigned one-to-one. We built such
items and rescored them. On the 500-feature runs the centre and count rules fall from 0.979 and 0.969 to
0.494 and 0.499. The stored scorer's weights, llama-3.1-70b \citep{grattafiori2024llama3} run locally (0.861 on the
original items; stored 0.812), change by $-0.001$ [$-0.005$, 0.003]. On the Gemma Scope 9B pair of \cref{sec:harm} the
131k minus 16k gap changes by $+0.004$ [$-0.007$, 0.015], inside the published 0.03. Because gemini-2.0-flash is retired,
gemma-3-27b-it \citep{gemmateam2025gemma3} in its place rises by 0.028, its gap change not resolved (\cref{app:dicts}).

\textbf{Random-offset windows remove the position cue; stitching only moves it.} Stitching four stored snippets per
passage puts the peaks at the piece centres. A prior there, recovered by replaying the stitching script, scores 0.496,
and Jev's 0.476 is not above it (26 of 49 features, $p=0.78$; \cref{app:stitch}). For 49 of the 50 features we therefore computed activations
locally on OpenWebText \citep{gokaslan2019openwebtext} with GPT-2 small and the same SAE. We
cut 64-token windows around five activating occurrences per feature with the peak offset drawn uniformly,
plus five random windows; 488 windows in total (one feature has eight). The preregistered bound was mean
$|r|\le0.05$ for the centre and four-quarter priors. The centre prior scores 0.003 and the four-quarter
prior $-0.005$ (\cref{fig:results}b). The best offset shared by all features, chosen after the fact over
the 64 offsets (index 16), scores 0.015$^\dagger$. An oracle that picks the best offset separately for each
feature using the true activations reaches 0.171. Because the set mixes activating and random windows,
the sequence cue remains ($\eta=0.166$); we report $r_w$ beside it rather than remove it. A first build was
discarded before scoring (\cref{app:prereg}).

\textbf{Activations need the BOS token.} Without it, activations recomputed locally on the stored crops match
Neuronpedia's at a median per-feature correlation of only 0.082; with it, 0.978 on the crops and 0.999 on the
stitched passages. Without it the first real token's residual, an outlier \citep{xiao2023sinks,sun2024massive,voita2023neurons},
has 29.4 times the median norm of the others and receives a median 84\% of the SAE's activation mass (\cref{app:prereg}).

\textbf{On clean windows, Jev reads the explanation.} Jev's architecture and training data are not
disclosed \citep{typesafe2026jev}. It is an
existence proof that the standard can be met, not a recommended scorer: on the simulation protocol the two open typed decision
models we tried did not beat their shuffled controls (\cref{tab:laya}). Each token is one question over an ordered
10-level rubric, read out as the expected level (\cref{app:protocol}). On the
natural windows it reaches a pooled correlation of 0.331 over 49 features with the 10-level rubric, against 0.013
with a shuffled explanation (46 of 49 features, $p=7.0\times10^{-11}$; \cref{fig:cases}). It is also above the
per-feature offset oracle (38 of 49, $p=1.4\times10^{-4}$)
and, before multiplicity correction, above the sequence-mean oracle ($\eta=0.166$; 34 of 49, $p=0.0094$). By \cref{prop:affine}(b)
its score splits into 0.285 within sequences and 0.045 between them. Most comes from ranking tokens within a window, where shared position priors score near zero.

\textbf{An open 8B simulator also reads it, at 59\% of Jev's level.} Llama-3.1-8B-Instruct with the original
all-at-once read-out on identical inputs also reads the explanation on these windows (0.194 against 0.019
shuffled; 41 of 49, $p=2.0\times10^{-6}$), at 59\% of Jev's level. It is not distinguishable from the
per-feature offset oracle (25 of 49), and on the stored snippets it, like Jev, scores below the centre prior
(0.386 against 0.720; \cref{app:protocol}).

\subsection{How does Jev compare with GPT-4's simulator?}\label{sec:h2h}

We ran Jev (10-level rubric) on the 400 neurons with GPT-4's explanations and sequences, with the primary test and its outcome wording fixed after a 40-neuron pilot. Jev's training data are not disclosed, so OpenAI's release (public
since 2023) may be among them; the natural windows of \cref{sec:clean} and their activations are new; their text
is public OpenWebText.

\textbf{Jev's pooled lead over GPT-4's simulator lies mostly between sequences.} Jev's mean pooled correlation is
0.197 against GPT-4's 0.160, a paired difference of
$+0.0373\pm0.0061$; Jev is higher on 256 of 400 neurons ($p=2.3\times10^{-8}$), the preregistered
``significantly higher'' outcome. Shuffled explanations drop Jev to 0.017 (334 wins to 66,
$p=3.3\times10^{-44}$). In the split of \cref{prop:affine}(b)$^\dagger$, the within-sequence term carries
$+0.010$ of the $+0.037$ gap and the between-sequence term $+0.027$.

\textbf{The lead is not detected in rank correlation, and both score below $\eta$.} Mean Spearman $\rho$ is 0.074 for Jev and 0.083 for GPT-4
($p=0.051$). For the 4-level rubric the preregistered adverse branch holds: GPT-4 is higher in rank correlation
($-0.016$, $p=0.014$). Jev sees the whole sequence, GPT-4 the
prefix (restricted to it, Jev changes by $-0.0023$, an order of magnitude below the gap; \cref{app:h2h}). Both score below $\eta$, so here a pooled score alone cannot show token-level skill.

\section{Limitations}\label{sec:limits}

Fuzzing marks are rebuilt; they equal the pinned fork's, but Neuronpedia's deployed commit is unknown. Use of a cue by a scorer is measured with the run and the label held fixed but not the
item (\cref{sec:harm}), and cross-dictionary levels are rule levels, not scores. One SAE carries stored runs
at scale, the Gemma result rests on 27 runs of 12 features, and delphi items are built locally (with a
BOS token delphi omits). The harm test (one scorer, 49 features,
one SAE) fired in 0 of 200 draws with a centre cue on half of 43 features (\cref{app:protocol}); ``no ranking harm
shown'' is not ``no ranking harm''. The matched-mark rescoring covers one stored scorer's weights and one dictionary
pair. The simulator comparison covers GPT-2 XL neurons, GPT-4 as released in 2023 and one open 8B simulator;
no current closed LLM simulator was run. All Jev results were requested as
\texttt{jev-1.13.0} with the exact rubric wording; the API is close to, not exactly, deterministic
(\cref{app:protocol}).
The count rule is told the negatives' mark count.

\section{Conclusion}\label{sec:conclusion}

 A rule told the negatives' mark count
beats the stored LLM scorers on every stored fuzzing set we measured; the seed-free count rule reaches 0.969 against
their 0.838. On the data tested, one of two stored
LLM scorers accepts centre-unmarked positives less often (the item not held fixed). Jev's levels are higher on stored
snippets than on natural windows (sets that also differ in length and composition; the gap follows length, not
centring), and the cue is not shown to reorder features. Removing the mark cues moves neither the stored llama
scorer's level nor the tested dictionary gap. Benchmarks should run four checks: report the best
explanation-free rule beside every score, cut windows at random offsets with the BOS
token, mark negatives from the positives' law, and report simulation scores' within-sequence part. On
such data a scorer can still show it reads the explanation. The results do not show that published orderings of
features or dictionaries are wrong. 

\section*{Impact Statement}
This paper concerns the reliability of automated-interpretability evaluation. We analyse public evaluation
data, public models and a commercial API; no human subjects or personal data are involved, and the
web-text snippets may contain offensive content. The shortcuts we describe could be used to inflate scores;
we report ways to remove or separate them. Those affected are authors and readers of SAE and
interpretability papers that rank dictionaries or explainers by these scores, and anyone relying on such scores
for safety claims. The positive impact is fewer inflated interpretability claims; the four checks in
\cref{sec:conclusion} are the mitigation for the misuse noted above.
\ifnum\arxivversion=1
\paragraph{Competing interests.} The authors have no affiliation with TypeSafe AI and received no funding from it; Jev was accessed through its public API.
\fi

\bibliographystyle{icml2026}
\bibliography{refs}

\begin{thebibliography}{70}
\providecommand{\natexlab}[1]{#1}
\providecommand{\url}[1]{\texttt{#1}}
\expandafter\ifx\csname urlstyle\endcsname\relax
  \providecommand{\doi}[1]{doi: #1}\else
  \providecommand{\doi}{doi: \begingroup \urlstyle{rm}\Url}\fi

\bibitem[Adebayo et~al.(2018)Adebayo, Gilmer, Muelly, Goodfellow, Hardt, and
  Kim]{adebayo2018sanity}
Adebayo, J., Gilmer, J., Muelly, M., Goodfellow, I., Hardt, M., and Kim, B.
\newblock {Sanity Checks for Saliency Maps}.
\newblock In \emph{Advances in Neural Information Processing Systems
  (NeurIPS)}, volume~31, 2018.

\bibitem[Almeida(2026)]{typesafe2026jev}
Almeida, D.
\newblock {Introducing System One Models \& Jev}.
\newblock TypeSafe AI Blog,
  \url{https://typesafe.ai/blog/introducing-system-one-models-and-jev}, 2026.
\newblock 15 September 2026.

\bibitem[Bau et~al.(2017)Bau, Zhou, Khosla, Oliva, and
  Torralba]{bau2017dissection}
Bau, D., Zhou, B., Khosla, A., Oliva, A., and Torralba, A.
\newblock {Network Dissection: Quantifying Interpretability of Deep Visual
  Representations}.
\newblock In \emph{IEEE Conference on Computer Vision and Pattern Recognition
  (CVPR)}, pp.\  3319--3327, 2017.
\newblock \doi{10.1109/CVPR.2017.354}.

\bibitem[Bills et~al.(2023)Bills, Cammarata, Mossing, Tillman, Gao, Goh,
  Sutskever, Leike, Wu, and Saunders]{bills2023neurons}
Bills, S., Cammarata, N., Mossing, D., Tillman, H., Gao, L., Goh, G.,
  Sutskever, I., Leike, J., Wu, J., and Saunders, W.
\newblock Language models can explain neurons in language models.
\newblock
  \url{https://openaipublic.blob.core.windows.net/neuron-explainer/paper/index.html},
  2023.

\bibitem[Bloom(2024)]{bloom2024gpt2saes}
Bloom, J.
\newblock {GPT-2 Small} residual-stream sparse autoencoders.
\newblock \url{https://huggingface.co/jbloom/GPT2-Small-SAEs-Reformatted},
  2024.

\bibitem[Bolukbasi et~al.(2021)Bolukbasi, Pearce, Yuan, Coenen, Reif, Viégas,
  and Wattenberg]{bolukbasi2021illusion}
Bolukbasi, T., Pearce, A., Yuan, A., Coenen, A., Reif, E., Viégas, F., and
  Wattenberg, M.
\newblock {An Interpretability Illusion for BERT}.
\newblock arXiv:2104.07143, 2021.

\bibitem[Bricken et~al.(2023)Bricken, Templeton, Batson, Chen,
  et~al.]{bricken2023monosemanticity}
Bricken, T., Templeton, A., Batson, J., Chen, B., et~al.
\newblock Towards monosemanticity: Decomposing language models with dictionary
  learning.
\newblock Transformer Circuits Thread,
  \url{https://transformer-circuits.pub/2023/monosemantic-features}, 2023.

\bibitem[Chen et~al.(2024)Chen, Zaharia, and Zou]{chen2023frugalgpt}
Chen, L., Zaharia, M., and Zou, J.
\newblock {FrugalGPT: How to Use Large Language Models While Reducing Cost and
  Improving Performance}.
\newblock \emph{Transactions on Machine Learning Research (TMLR)}, 2024.

\bibitem[Choi et~al.(2024)Choi, Huang, Meng, Johnson, Steinhardt, and
  Schwettmann]{choi2024scaling}
Choi, D., Huang, V., Meng, K., Johnson, D.~D., Steinhardt, J., and Schwettmann,
  S.
\newblock Scaling automatic neuron description.
\newblock Transluce report, \url{https://transluce.org/neuron-descriptions},
  2024.

\bibitem[{Convai Innovations}(2026)]{laya2026}
{Convai Innovations}.
\newblock {Laya: Multilingual, Non-Autoregressive System 1 Decision Engine}.
\newblock GitHub repository, \url{https://github.com/NandhaKishorM/laya};
  models at \url{https://huggingface.co/convaiinnovations}, 2026.

\bibitem[Cunningham et~al.(2024)Cunningham, Ewart, Riggs, Huben, and
  Sharkey]{cunningham2023sae}
Cunningham, H., Ewart, A., Riggs, L., Huben, R., and Sharkey, L.
\newblock {Sparse Autoencoders Find Highly Interpretable Features in Language
  Models}.
\newblock In \emph{International Conference on Learning Representations
  (ICLR)}, 2024.

\bibitem[Devlin et~al.(2019)Devlin, Chang, Lee, and Toutanova]{devlin2019bert}
Devlin, J., Chang, M.-W., Lee, K., and Toutanova, K.
\newblock {BERT: Pre-training of Deep Bidirectional Transformers for Language
  Understanding}.
\newblock In \emph{Proceedings of the 2019 Conference of the North American
  Chapter of the Association for Computational Linguistics: Human Language
  Technologies (NAACL-HLT), Volume 1 (Long and Short Papers)}, pp.\
  4171--4186. Association for Computational Linguistics, 2019.
\newblock \doi{10.18653/v1/N19-1423}.

\bibitem[Ding et~al.(2024)Ding, Mallick, Wang, Sim, Mukherjee, Ruhle,
  Lakshmanan, and Awadallah]{ding2024hybrid}
Ding, D., Mallick, A., Wang, C., Sim, R., Mukherjee, S., Ruhle, V., Lakshmanan,
  L. V.~S., and Awadallah, A.~H.
\newblock {Hybrid LLM: Cost-Efficient and Quality-Aware Query Routing}.
\newblock In \emph{International Conference on Learning Representations
  (ICLR)}, 2024.

\bibitem[{EleutherAI}(2024)]{eleuther2024delphi}
{EleutherAI}.
\newblock {delphi}.
\newblock GitHub repository, \url{https://github.com/EleutherAI/delphi}, 2024.

\bibitem[Gao et~al.(2025)Gao, la~Tour, Tillman, Goh, Troll, Radford, Sutskever,
  Leike, and Wu]{gao2024scaling}
Gao, L., la~Tour, T.~D., Tillman, H., Goh, G., Troll, R., Radford, A.,
  Sutskever, I., Leike, J., and Wu, J.
\newblock {Scaling and Evaluating Sparse Autoencoders}.
\newblock In \emph{International Conference on Learning Representations
  (ICLR)}, 2025.
\newblock URL \url{https://openreview.net/forum?id=tcsZt9ZNKD}.

\bibitem[Geirhos et~al.(2020)Geirhos, Jacobsen, Michaelis, Zemel, Brendel,
  Bethge, and Wichmann]{geirhos2020shortcut}
Geirhos, R., Jacobsen, J.-H., Michaelis, C., Zemel, R., Brendel, W., Bethge,
  M., and Wichmann, F.~A.
\newblock {Shortcut learning in deep neural networks}.
\newblock \emph{Nature Machine Intelligence}, 2\penalty0 (11):\penalty0
  665--673, 2020.
\newblock \doi{10.1038/s42256-020-00257-z}.

\bibitem[{Gemini Team}(2024)]{geminiteam2024gemini15}
{Gemini Team}.
\newblock {Gemini 1.5: Unlocking Multimodal Understanding across Millions of
  Tokens of Context}.
\newblock arXiv:2403.05530, 2024.

\bibitem[{Gemma Team}(2024)]{gemmateam2024gemma2}
{Gemma Team}.
\newblock {Gemma 2: Improving Open Language Models at a Practical Size}.
\newblock arXiv:2408.00118, 2024.

\bibitem[{Gemma Team}(2025)]{gemmateam2025gemma3}
{Gemma Team}.
\newblock {Gemma 3 Technical Report}.
\newblock arXiv:2503.19786, 2025.

\bibitem[Gilardi et~al.(2023)Gilardi, Alizadeh, and Kubli]{gilardi2023chatgpt}
Gilardi, F., Alizadeh, M., and Kubli, M.
\newblock {ChatGPT outperforms crowd workers for text-annotation tasks}.
\newblock \emph{Proceedings of the National Academy of Sciences}, 120\penalty0
  (30):\penalty0 e2305016120, 2023.
\newblock \doi{10.1073/pnas.2305016120}.

\bibitem[Gokaslan \& Cohen(2019)Gokaslan and Cohen]{gokaslan2019openwebtext}
Gokaslan, A. and Cohen, V.
\newblock {OpenWebText Corpus}.
\newblock \url{http://Skylion007.github.io/OpenWebTextCorpus}, 2019.

\bibitem[Grattafiori et~al.(2024)Grattafiori, Dubey, Jauhri, Pandey,
  et~al.]{grattafiori2024llama3}
Grattafiori, A., Dubey, A., Jauhri, A., Pandey, A., et~al.
\newblock {The Llama 3 Herd of Models}.
\newblock arXiv:2407.21783, 2024.

\bibitem[Guo et~al.(2017)Guo, Pleiss, Sun, and Weinberger]{guo2017calibration}
Guo, C., Pleiss, G., Sun, Y., and Weinberger, K.~Q.
\newblock {On Calibration of Modern Neural Networks}.
\newblock In \emph{Proceedings of the 34th International Conference on Machine
  Learning (ICML)}, volume~70 of \emph{Proceedings of Machine Learning
  Research}, pp.\  1321--1330, 2017.

\bibitem[Gur-Arieh et~al.(2025)Gur-Arieh, Mayan, Agassy, Geiger, and
  Geva]{gurarieh2025output}
Gur-Arieh, Y., Mayan, R., Agassy, C., Geiger, A., and Geva, M.
\newblock {Enhancing Automated Interpretability with Output-Centric Feature
  Descriptions}.
\newblock In \emph{Proceedings of the 63rd Annual Meeting of the Association
  for Computational Linguistics (ACL)}, pp.\  5757--5778. Association for
  Computational Linguistics, 2025.
\newblock \doi{10.18653/v1/2025.acl-long.288}.

\bibitem[Gururangan et~al.(2018)Gururangan, Swayamdipta, Levy, Schwartz,
  Bowman, and Smith]{gururangan2018annotation}
Gururangan, S., Swayamdipta, S., Levy, O., Schwartz, R., Bowman, S.~R., and
  Smith, N.~A.
\newblock {Annotation Artifacts in Natural Language Inference Data}.
\newblock In \emph{Proceedings of the 2018 Conference of the North American
  Chapter of the Association for Computational Linguistics: Human Language
  Technologies (NAACL-HLT), Volume 2 (Short Papers)}, pp.\  107--112.
  Association for Computational Linguistics, 2018.
\newblock \doi{10.18653/v1/N18-2017}.

\bibitem[Heap et~al.(2026)Heap, Lawson, Farnik, and Aitchison]{heap2025random}
Heap, T., Lawson, T., Farnik, L., and Aitchison, L.
\newblock {Automated Interpretability Metrics Do Not Distinguish Trained and
  Random Transformers}.
\newblock In \emph{International Conference on Learning Representations
  (ICLR)}, 2026.

\bibitem[Hernandez et~al.(2022)Hernandez, Schwettmann, Bau, Bagashvili,
  Torralba, and Andreas]{hernandez2022milan}
Hernandez, E., Schwettmann, S., Bau, D., Bagashvili, T., Torralba, A., and
  Andreas, J.
\newblock {Natural Language Descriptions of Deep Visual Features}.
\newblock In \emph{International Conference on Learning Representations
  (ICLR)}, 2022.

\bibitem[Huang et~al.(2023)Huang, Geiger, D'Oosterlinck, Wu, and
  Potts]{huang2023rigorous}
Huang, J., Geiger, A., D'Oosterlinck, K., Wu, Z., and Potts, C.
\newblock {Rigorously Assessing Natural Language Explanations of Neurons}.
\newblock In \emph{Proceedings of the 6th BlackboxNLP Workshop: Analyzing and
  Interpreting Neural Networks for NLP}, pp.\  317--331. Association for
  Computational Linguistics, 2023.
\newblock \doi{10.18653/v1/2023.blackboxnlp-1.24}.

\bibitem[Ibrahim \& Zaki(2026)Ibrahim and Zaki]{ibrahim2026decision}
Ibrahim, H. and Zaki, Y.
\newblock {Evaluating Decision Models for Text Annotation in Computational
  Social Science}.
\newblock arXiv:2609.24574, 2026.

\bibitem[Juang et~al.(2024)Juang, Paulo, Drori, and
  Belrose]{eleuther2024autointerp}
Juang, C., Paulo, G., Drori, J., and Belrose, N.
\newblock Open source automated interpretability for sparse autoencoder
  features.
\newblock EleutherAI Blog, \url{https://blog.eleuther.ai/autointerp/}, 2024.

\bibitem[Kadavath et~al.(2022)Kadavath, Conerly, Askell, Henighan, Drain,
  Perez, Schiefer, Hatfield-Dodds, DasSarma, Tran-Johnson, Johnston, El-Showk,
  Jones, Elhage, Hume, Chen, Bai, Bowman, Fort, Ganguli, Hernandez, Jacobson,
  Kernion, Kravec, Lovitt, Ndousse, Olsson, Ringer, Amodei, Brown, Clark,
  Joseph, Mann, McCandlish, Olah, and Kaplan]{kadavath2022know}
Kadavath, S., Conerly, T., Askell, A., Henighan, T., Drain, D., Perez, E.,
  Schiefer, N., Hatfield-Dodds, Z., DasSarma, N., Tran-Johnson, E., Johnston,
  S., El-Showk, S., Jones, A., Elhage, N., Hume, T., Chen, A., Bai, Y., Bowman,
  S., Fort, S., Ganguli, D., Hernandez, D., Jacobson, J., Kernion, J., Kravec,
  S., Lovitt, L., Ndousse, K., Olsson, C., Ringer, S., Amodei, D., Brown, T.,
  Clark, J., Joseph, N., Mann, B., McCandlish, S., Olah, C., and Kaplan, J.
\newblock {Language Models (Mostly) Know What They Know}.
\newblock arXiv:2207.05221, 2022.

\bibitem[Kantamneni et~al.(2025)Kantamneni, Engels, Rajamanoharan, Tegmark, and
  Nanda]{kantamneni2025probing}
Kantamneni, S., Engels, J., Rajamanoharan, S., Tegmark, M., and Nanda, N.
\newblock {Are Sparse Autoencoders Useful? A Case Study in Sparse Probing}.
\newblock In \emph{Proceedings of the 42nd International Conference on Machine
  Learning (ICML)}, volume 267 of \emph{Proceedings of Machine Learning
  Research}, pp.\  29018--29049, 2025.

\bibitem[Kapoor \& Narayanan(2023)Kapoor and Narayanan]{kapoor2023leakage}
Kapoor, S. and Narayanan, A.
\newblock {Leakage and the reproducibility crisis in machine-learning-based
  science}.
\newblock \emph{Patterns}, 4\penalty0 (9):\penalty0 100804, 2023.
\newblock \doi{10.1016/j.patter.2023.100804}.

\bibitem[Karvonen et~al.(2025)Karvonen, Rager, Lin, Tigges, Bloom, Chanin, Lau,
  Farrell, McDougall, Ayonrinde, Till, Wearden, Conmy, Marks, and
  Nanda]{karvonen2025saebench}
Karvonen, A., Rager, C., Lin, J., Tigges, C., Bloom, J., Chanin, D., Lau,
  Y.-T., Farrell, E., McDougall, C., Ayonrinde, K., Till, D., Wearden, M.,
  Conmy, A., Marks, S., and Nanda, N.
\newblock {SAEBench: A Comprehensive Benchmark for Sparse Autoencoders in
  Language Model Interpretability}.
\newblock In \emph{Proceedings of the 42nd International Conference on Machine
  Learning (ICML)}, volume 267 of \emph{Proceedings of Machine Learning
  Research}, pp.\  29223--29264, 2025.

\bibitem[Kopf et~al.(2024)Kopf, Bommer, Hedström, Lapuschkin, Höhne, and
  Bykov]{kopf2024cosy}
Kopf, L., Bommer, P.~L., Hedström, A., Lapuschkin, S., Höhne, M. M.~C., and
  Bykov, K.
\newblock {CoSy: Evaluating Textual Explanations of Neurons}.
\newblock In \emph{Advances in Neural Information Processing Systems
  (NeurIPS)}, volume~37, pp.\  34656--34685, 2024.
\newblock \doi{10.52202/079017-1093}.

\bibitem[Korznikov et~al.(2026)Korznikov, Galichin, Dontsov, Rogov, Oseledets,
  and Tutubalina]{korznikov2026sanity}
Korznikov, A., Galichin, A., Dontsov, A., Rogov, O., Oseledets, I., and
  Tutubalina, E.
\newblock {Sanity Checks for Sparse Autoencoders: Do SAEs Beat Random
  Baselines?}
\newblock arXiv:2602.14111, 2026.

\bibitem[Lehmann \& Romano(2005)Lehmann and Romano]{lehmann2005testing}
Lehmann, E.~L. and Romano, J.~P.
\newblock \emph{Testing Statistical Hypotheses}.
\newblock Springer Texts in Statistics. Springer, New York, NY, third edition,
  2005.
\newblock \doi{10.1007/0-387-27605-X}.

\bibitem[Lieberum et~al.(2024)Lieberum, Rajamanoharan, Conmy, Smith, Sonnerat,
  Varma, Kramár, Dragan, Shah, and Nanda]{lieberum2024gemmascope}
Lieberum, T., Rajamanoharan, S., Conmy, A., Smith, L., Sonnerat, N., Varma, V.,
  Kramár, J., Dragan, A., Shah, R., and Nanda, N.
\newblock {Gemma Scope: Open Sparse Autoencoders Everywhere All At Once on
  Gemma 2}.
\newblock In \emph{Proceedings of the 7th BlackboxNLP Workshop: Analyzing and
  Interpreting Neural Networks for NLP}, pp.\  278--300. Association for
  Computational Linguistics, 2024.
\newblock \doi{10.18653/v1/2024.blackboxnlp-1.19}.

\bibitem[Lin(2023)]{neuronpedia}
Lin, J.
\newblock {Neuronpedia}: Interactive reference and tooling for analyzing neural
  networks.
\newblock \url{https://www.neuronpedia.org}, 2023.

\bibitem[Makelov et~al.(2025)Makelov, Lange, and Nanda]{makelov2024principled}
Makelov, A., Lange, G., and Nanda, N.
\newblock {Towards Principled Evaluations of Sparse Autoencoders for
  Interpretability and Control}.
\newblock In \emph{International Conference on Learning Representations
  (ICLR)}, 2025.

\bibitem[Marone et~al.(2026)Marone, Weller, Fleshman, Yang, Lawrie, and {Van
  Durme}]{marone2025mmbert}
Marone, M., Weller, O., Fleshman, W., Yang, E., Lawrie, D., and {Van Durme}, B.
\newblock {mmBERT: A Modern Multilingual Encoder with Annealed Language
  Learning}.
\newblock In \emph{International Conference on Machine Learning (ICML)}, 2026.

\bibitem[McCoy et~al.(2019)McCoy, Pavlick, and Linzen]{mccoy2019right}
McCoy, R.~T., Pavlick, E., and Linzen, T.
\newblock {Right for the Wrong Reasons: Diagnosing Syntactic Heuristics in
  Natural Language Inference}.
\newblock In \emph{Proceedings of the 57th Annual Meeting of the Association
  for Computational Linguistics (ACL)}, pp.\  3428--3448. Association for
  Computational Linguistics, 2019.
\newblock \doi{10.18653/v1/P19-1334}.

\bibitem[Oikarinen \& Weng(2023)Oikarinen and Weng]{oikarinen2022clipdissect}
Oikarinen, T. and Weng, T.-W.
\newblock {CLIP-Dissect: Automatic Description of Neuron Representations in
  Deep Vision Networks}.
\newblock In \emph{International Conference on Learning Representations
  (ICLR)}, 2023.

\bibitem[Oikarinen et~al.(2025)Oikarinen, Yan, and Weng]{oikarinen2025unified}
Oikarinen, T., Yan, G., and Weng, T.-W.
\newblock {Evaluating Neuron Explanations: A Unified Framework with Sanity
  Checks}.
\newblock In \emph{Proceedings of the 42nd International Conference on Machine
  Learning (ICML)}, volume 267 of \emph{Proceedings of Machine Learning
  Research}, pp.\  47090--47131, 2025.

\bibitem[Ong et~al.(2025)Ong, Almahairi, Wu, Chiang, Wu, Gonzalez, Kadous, and
  Stoica]{ong2024routellm}
Ong, I., Almahairi, A., Wu, V., Chiang, W.-L., Wu, T., Gonzalez, J.~E., Kadous,
  M.~W., and Stoica, I.
\newblock {RouteLLM: Learning to Route LLMs from Preference Data}.
\newblock In \emph{International Conference on Learning Representations
  (ICLR)}, 2025.

\bibitem[{OpenAI}(2023)]{openai2023gpt4}
{OpenAI}.
\newblock {GPT-4 Technical Report}.
\newblock arXiv:2303.08774, 2023.

\bibitem[Paulo \& Belrose(2026)Paulo and Belrose]{paulo2025noexplanations}
Paulo, G. and Belrose, N.
\newblock {Evaluating SAE interpretability without generating explanations}.
\newblock In \emph{International Conference on Learning Representations
  (ICLR)}, 2026.

\bibitem[Paulo et~al.(2025)Paulo, Mallen, Juang, and
  Belrose]{paulo2024autointerp}
Paulo, G., Mallen, A., Juang, C., and Belrose, N.
\newblock {Automatically Interpreting Millions of Features in Large Language
  Models}.
\newblock In \emph{Proceedings of the 42nd International Conference on Machine
  Learning (ICML)}, volume 267 of \emph{Proceedings of Machine Learning
  Research}, pp.\  48393--48421, 2025.

\bibitem[Poliak et~al.(2018)Poliak, Naradowsky, Haldar, Rudinger, and {Van
  Durme}]{poliak2018hypothesis}
Poliak, A., Naradowsky, J., Haldar, A., Rudinger, R., and {Van Durme}, B.
\newblock {Hypothesis Only Baselines in Natural Language Inference}.
\newblock In \emph{Proceedings of the Seventh Joint Conference on Lexical and
  Computational Semantics (*SEM)}, pp.\  180--191. Association for
  Computational Linguistics, 2018.
\newblock \doi{10.18653/v1/S18-2023}.

\bibitem[Puri et~al.(2025)Puri, Jain, Golimblevskaia, Kahardipraja, Wiegand,
  Samek, and Lapuschkin]{puri2025fade}
Puri, B., Jain, A., Golimblevskaia, E., Kahardipraja, P., Wiegand, T., Samek,
  W., and Lapuschkin, S.
\newblock {FADE: Why Bad Descriptions Happen to Good Features}.
\newblock In \emph{Findings of the Association for Computational Linguistics:
  ACL 2025}, pp.\  17138--17160. Association for Computational Linguistics,
  2025.
\newblock \doi{10.18653/v1/2025.findings-acl.881}.

\bibitem[Radford et~al.(2019)Radford, Wu, Child, Luan, Amodei, and
  Sutskever]{radford2019gpt2}
Radford, A., Wu, J., Child, R., Luan, D., Amodei, D., and Sutskever, I.
\newblock Language models are unsupervised multitask learners.
\newblock OpenAI technical report, 2019.
\newblock URL
  \url{https://cdn.openai.com/better-language-models/language_models_are_unsupervised_multitask_learners.pdf}.

\bibitem[Robinson(1950)]{robinson1950ecological}
Robinson, W.~S.
\newblock Ecological correlations and the behavior of individuals.
\newblock \emph{American Sociological Review}, 15\penalty0 (3):\penalty0
  351--357, 1950.
\newblock \doi{10.2307/2087176}.

\bibitem[Schick \& Schütze(2021)Schick and Schütze]{schick2021pet}
Schick, T. and Schütze, H.
\newblock {Exploiting Cloze-Questions for Few-Shot Text Classification and
  Natural Language Inference}.
\newblock In \emph{Proceedings of the 16th Conference of the European Chapter
  of the Association for Computational Linguistics (EACL): Main Volume}, pp.\
  255--269. Association for Computational Linguistics, 2021.
\newblock \doi{10.18653/v1/2021.eacl-main.20}.

\bibitem[Schwettmann et~al.(2023)Schwettmann, Shaham, Materzynska, Chowdhury,
  Li, Andreas, Bau, and Torralba]{schwettmann2023find}
Schwettmann, S., Shaham, T.~R., Materzynska, J., Chowdhury, N., Li, S.,
  Andreas, J., Bau, D., and Torralba, A.
\newblock {FIND: A Function Description Benchmark for Evaluating
  Interpretability Methods}.
\newblock In \emph{Advances in Neural Information Processing Systems (NeurIPS),
  Datasets and Benchmarks Track}, volume~36, 2023.
\newblock \doi{10.52202/075280-3308}.

\bibitem[Shaham et~al.(2024)Shaham, Schwettmann, Wang, Rajaram, Hernandez,
  Andreas, and Torralba]{shaham2024maia}
Shaham, T.~R., Schwettmann, S., Wang, F., Rajaram, A., Hernandez, E., Andreas,
  J., and Torralba, A.
\newblock {A Multimodal Automated Interpretability Agent}.
\newblock In \emph{Proceedings of the 41st International Conference on Machine
  Learning (ICML)}, volume 235 of \emph{Proceedings of Machine Learning
  Research}, pp.\  44293--44321, 2024.

\bibitem[Singh et~al.(2023)Singh, Hsu, Antonello, Jain, Huth, Yu, and
  Gao]{singh2023sasc}
Singh, C., Hsu, A.~R., Antonello, R., Jain, S., Huth, A.~G., Yu, B., and Gao,
  J.
\newblock {Explaining black box text modules in natural language with language
  models}.
\newblock arXiv:2305.09863, 2023.

\bibitem[Sun et~al.(2024)Sun, Chen, Kolter, and Liu]{sun2024massive}
Sun, M., Chen, X., Kolter, J.~Z., and Liu, Z.
\newblock {Massive Activations in Large Language Models}.
\newblock In \emph{Conference on Language Modeling (COLM)}, 2024.

\bibitem[Sun et~al.(2026)Sun, Xu, Shi, and Yang]{sun2026typesafe}
Sun, Y., Xu, J., Shi, J., and Yang, Z.
\newblock {Type-Safe Is Not Error-Free: Typed Decision Models Follow the Option
  Name, Not the Definition Bound to It}.
\newblock arXiv:2609.26758, 2026.

\bibitem[Tate(1954)]{tate1954pointbiserial}
Tate, R.~F.
\newblock Correlation between a discrete and a continuous variable.
  {Point-Biserial} correlation.
\newblock \emph{The Annals of Mathematical Statistics}, 25\penalty0
  (3):\penalty0 603--607, 1954.
\newblock \doi{10.1214/aoms/1177728730}.

\bibitem[Templeton et~al.(2024)Templeton, Conerly, Marcus, Lindsey, Bricken,
  et~al.]{templeton2024scaling}
Templeton, A., Conerly, T., Marcus, J., Lindsey, J., Bricken, T., et~al.
\newblock Scaling monosemanticity: Extracting interpretable features from
  {C}laude 3 {S}onnet.
\newblock Transformer Circuits Thread,
  \url{https://transformer-circuits.pub/2024/scaling-monosemanticity}, 2024.

\bibitem[van~der Ben et~al.(2026)van~der Ben, Roch, Hedstr{\"o}m, and
  El-Assady]{vanderben2026building}
van~der Ben, S., Roch, N., Hedstr{\"o}m, A., and El-Assady, M.
\newblock {Building Fast, Evaluating Slow: Pipeline Choices Dominate
  Autointerpretability Score Variance}.
\newblock arXiv:2607.19386, 2026.

\bibitem[Voita et~al.(2024)Voita, Ferrando, and Nalmpantis]{voita2023neurons}
Voita, E., Ferrando, J., and Nalmpantis, C.
\newblock {Neurons in Large Language Models: Dead, N-gram, Positional}.
\newblock In \emph{Findings of the Association for Computational Linguistics:
  ACL 2024}, pp.\  1288--1301. Association for Computational Linguistics, 2024.
\newblock \doi{10.18653/v1/2024.findings-acl.75}.

\bibitem[Wang et~al.(2024)Wang, Li, Chen, Cai, Zhu, Lin, Cao, Kong, Liu, Liu,
  and Sui]{wang2023fair}
Wang, P., Li, L., Chen, L., Cai, Z., Zhu, D., Lin, B., Cao, Y., Kong, L., Liu,
  Q., Liu, T., and Sui, Z.
\newblock {Large Language Models are not Fair Evaluators}.
\newblock In \emph{Proceedings of the 62nd Annual Meeting of the Association
  for Computational Linguistics (ACL)}, pp.\  9440--9450. Association for
  Computational Linguistics, 2024.
\newblock \doi{10.18653/v1/2024.acl-long.511}.

\bibitem[Warner et~al.(2025)Warner, Chaffin, Clavié, Weller, Hallström,
  Taghadouini, Gallagher, Biswas, Ladhak, Aarsen, Adams, Howard, and
  Poli]{warner2024modernbert}
Warner, B., Chaffin, A., Clavié, B., Weller, O., Hallström, O., Taghadouini,
  S., Gallagher, A., Biswas, R., Ladhak, F., Aarsen, T., Adams, G.~T., Howard,
  J., and Poli, I.
\newblock {Smarter, Better, Faster, Longer: A Modern Bidirectional Encoder for
  Fast, Memory Efficient, and Long Context Finetuning and Inference}.
\newblock In \emph{Proceedings of the 63rd Annual Meeting of the Association
  for Computational Linguistics (ACL)}, pp.\  2526--2547. Association for
  Computational Linguistics, 2025.
\newblock \doi{10.18653/v1/2025.acl-long.127}.

\bibitem[Xiao et~al.(2024)Xiao, Tian, Chen, Han, and Lewis]{xiao2023sinks}
Xiao, G., Tian, Y., Chen, B., Han, S., and Lewis, M.
\newblock {Efficient Streaming Language Models with Attention Sinks}.
\newblock In \emph{International Conference on Learning Representations
  (ICLR)}, 2024.

\bibitem[Yang et~al.(2025)Yang, Li, Yang, Zhang, Hui, et~al.]{yang2025qwen3}
Yang, A., Li, A., Yang, B., Zhang, B., Hui, B., et~al.
\newblock {Qwen3} technical report.
\newblock arXiv:2505.09388, 2025.

\bibitem[Zhang et~al.(2026)Zhang, Yang, Shi, Chen, Nie, Wan, Hong, Liu, and
  Wu]{zhang2026anyjev}
Zhang, J., Yang, T., Shi, Y., Chen, E., Nie, Z., Wan, K., Hong, L., Liu, N.,
  and Wu, L.
\newblock {AnyJev} technical report.
\newblock arXiv:2610.00831, 2026.

\bibitem[Zhao et~al.(2021)Zhao, Wallace, Feng, Klein, and
  Singh]{zhao2021calibrate}
Zhao, T.~Z., Wallace, E., Feng, S., Klein, D., and Singh, S.
\newblock {Calibrate Before Use: Improving Few-Shot Performance of Language
  Models}.
\newblock In \emph{Proceedings of the 38th International Conference on Machine
  Learning (ICML)}, volume 139 of \emph{Proceedings of Machine Learning
  Research}, pp.\  12697--12706, 2021.

\bibitem[Zheng et~al.(2023)Zheng, Chiang, Sheng, Zhuang, Wu, Zhuang, Lin, Li,
  Li, Xing, Zhang, Gonzalez, and Stoica]{zheng2023judging}
Zheng, L., Chiang, W.-L., Sheng, Y., Zhuang, S., Wu, Z., Zhuang, Y., Lin, Z.,
  Li, Z., Li, D., Xing, E.~P., Zhang, H., Gonzalez, J.~E., and Stoica, I.
\newblock {Judging LLM-as-a-Judge with MT-Bench and Chatbot Arena}.
\newblock In \emph{Advances in Neural Information Processing Systems (NeurIPS),
  Datasets and Benchmarks Track}, volume~36, 2023.
\newblock \doi{10.52202/075280-2020}.

\bibitem[Ziems et~al.(2024)Ziems, Held, Shaikh, Chen, Zhang, and
  Yang]{ziems2024css}
Ziems, C., Held, W., Shaikh, O., Chen, J., Zhang, Z., and Yang, D.
\newblock {Can Large Language Models Transform Computational Social Science?}
\newblock \emph{Computational Linguistics}, 50\penalty0 (1):\penalty0 237--291,
  2024.
\newblock \doi{10.1162/coli_a_00502}.

\end{thebibliography}

\newpage
\appendix
\onecolumn
\raggedbottom

\section{Reproducibility Statement and Use of AI Assistants}\label{app:statements}

\paragraph{Reproducibility statement.} Code, preregistrations with appended outcomes, a lab notebook, cached API responses, per-item results and
a manifest with the size and SHA-256 of every input file will be released upon publication. Every number, table and figure is generated by scripts from the saved
model outputs; \cref{app:provenance} lists the data sources. Neuronpedia's
snippet text is replaced by content hashes (its data licence is unstated) and can be re-fetched from its
public API. Compute and API cost are in \cref{app:protocol} and \cref{app:cost}.

\paragraph{Use of AI assistants.} AI assistants helped write some of the experiment code and check its correctness, helped collect
public data, helped check the correctness of citations, and helped polish the paper.

\section{Proofs}\label{app:proofs}

Write $\langle x,y\rangle=\sum_{s,t}(x_{st}-\bar x)(y_{st}-\bar y)$ and $\|x\|=\langle x,x\rangle^{1/2}$,
so $r(x,y)=\langle x,y\rangle/(\|x\|\|y\|)$ whenever both norms are positive.

\paragraph{\Cref{prop:affine}.} (a) $\alpha\hat a+\beta$ has pooled mean $\alpha\bar{\hat a}+\beta$, so
$\langle a,\alpha\hat a+\beta\rangle=\alpha\langle a,\hat a\rangle$ and $\|\alpha\hat a+\beta\|=\alpha\|\hat a\|$
for $\alpha>0$; the ratio is unchanged. An increasing map that is not affine can change $r$: with
$a=\hat a=(0,1,2)$ and $f(x)=x^2$, $r(a,\hat a)=1$ but $r(a,f(\hat a))=6/\sqrt{39}\approx0.961$.
(b) Write $a=w+g$ with $w_{st}=a_{st}-\bar a_s$ and $g_{st}=\bar a_s$, and likewise $\hat a=\hat w+\hat g$.
Since $\sum_t w_{st}=0$ for every $s$ and $g$ is constant within sequences, $\sum_{s,t}w_{st}(g_{st}-\bar g)=0$;
the same holds for every within/between pair, and $\bar w=0$, $\bar g=\bar a$. Hence
$\langle a,\hat a\rangle=\langle w,\hat w\rangle+\langle g,\hat g\rangle$ and
$\|a\|^2=\|w\|^2+\|g\|^2$, so $\|g\|/\|a\|=\eta$ and $\|w\|/\|a\|=\sqrt{1-\eta^2}$, and likewise with
$\hat\eta$ for $\hat a$. Dividing by $\|a\|\|\hat a\|$,
\[
r(a,\hat a)=\frac{\langle w,\hat w\rangle}{\|w\|\|\hat w\|}\frac{\|w\|}{\|a\|}\frac{\|\hat w\|}{\|\hat a\|}
+\frac{\langle g,\hat g\rangle}{\|g\|\|\hat g\|}\frac{\|g\|}{\|a\|}\frac{\|\hat g\|}{\|\hat a\|}
=r_w\sqrt{1-\eta^2}\sqrt{1-\hat\eta^2}+r_b\,\eta\,\hat\eta ,
\]
where $r_b=\langle g,\hat g\rangle/(\|g\|\|\hat g\|)=\sum_s n_s(\bar a_s-\bar a)(\bar{\hat a}_s-\bar{\hat a})\big/
\big[\sum_s n_s(\bar a_s-\bar a)^2\sum_s n_s(\bar{\hat a}_s-\bar{\hat a})^2\big]^{1/2}$ is the $n_s$-weighted
correlation of sequence means (a term with a vanishing norm is zero on both sides). As $|r_b|\le1$ and
$\hat\eta\le1$, the second term is at most $\eta$.
(c) If $\hat a$ is constant within sequences, $\hat w=0$ and $\hat\eta=1$, so by (b) $r(a,\hat a)=r_b\eta\le\eta$,
with equality, when $\eta>0$, if and only if $r_b=1$, i.e.\ the sequence values $\bar{\hat a}_s$ are a
positive affine function of $\bar a_s$ over the sequences. The predictor $\hat a=g$ is such a function,
so it scores $\eta$. (If $\eta=0$, every such predictor scores $0=\eta$, and $g$ is constant.)
For a non-constant sequence-level offset $\delta$ (constant within sequences) and $\lambda>0$,
\[
r(a,\hat a+\lambda\delta)=\frac{\langle a,\hat a\rangle/\lambda+\langle a,\delta\rangle}
{\|a\|\,\big(\|\hat a\|^2/\lambda^2+2\langle\hat a,\delta\rangle/\lambda+\|\delta\|^2\big)^{1/2}}
\;\xrightarrow[\lambda\to\infty]{}\;\frac{\langle a,\delta\rangle}{\|a\|\,\|\delta\|}=r(a,\delta)\le\eta,
\]
the last step by the first sentence of (c) applied to $\delta$. $\square$

\paragraph{\Cref{prop:centre}.} Treat the $N$ pooled tokens as a population with uniform weight. Then
$\mathbb E[c]=\pi$, $\operatorname{Var}(c)=\pi(1-\pi)$, $\mathbb E[a]=\pi\mu_c+(1-\pi)\mu_o$, and
$\mathbb E[ac]=\pi\mu_c$, so
$\operatorname{Cov}(a,c)=\pi\mu_c-\pi(\pi\mu_c+(1-\pi)\mu_o)=\pi(1-\pi)(\mu_c-\mu_o)$. Dividing by
$\sigma_a\sqrt{\pi(1-\pi)}$ gives the first claim; all moments here carry weight $1/N$, so $\sigma_a$
must be the population standard deviation for the formula to hold ($\pi<1$ because some $n_s\ge2$). If $a_{st}=0$ off the centre, $\mu_o=0$; with $s_c^2$ the population variance (weight $1/m$) of
the $m$ centre activations, $\mathbb E[a^2]=\pi(\mu_c^2+s_c^2)$ and
$\sigma_a^2=\pi(\mu_c^2+s_c^2)-\pi^2\mu_c^2=\pi\mu_c^2(1-\pi+\kappa^2)$ with $\kappa=s_c/\mu_c$.
Substituting,
$r=\pi(1-\pi)\mu_c/\big(\sqrt{\pi(1-\pi)}\,\sqrt\pi\,\mu_c\sqrt{1-\pi+\kappa^2}\big)
=\sqrt{(1-\pi)/(1-\pi+\kappa^2)}$. $\square$

On the 50 stored-snippet features, $\pi=1/17$ for 47 features and slightly below it for the other 3,
whose sets include 6 snippets that are not 17 tokens long; the first identity reproduces the centre
prior's per-feature scores to $10^{-14}$ (an algebraic identity check on the same data; random-design checks
of all three propositions are in the released code).

\paragraph{\Cref{prop:marks}.} For a possibly randomised rule write $g(S)\in[0,1]$ for the probability that it answers ``yes'',
and $q_\pm=\mathbb E_\pm[g(S)]$. By \eqref{eq:ba}, $\mathrm{BA}(g)=\tfrac12(q_++1-q_-)=\tfrac12(1+q_+-q_-)$.
With $\nu$ dominating both laws, $q_+-q_-=\int g\,(p_+-p_-)\,d\nu\le\int\mathbf 1[p_+>p_-]\,(p_+-p_-)\,d\nu
=\mathrm{TV}(P_+,P_-)$, because $0\le g\le1$ and, as $\int(p_+-p_-)\,d\nu=0$,
$\int(p_+-p_-)^+d\nu=\tfrac12\int|p_+-p_-|\,d\nu$; the integrand is maximised pointwise by $g^\star$, which
attains the bound. If $P_+=P_-$, then $q_+=q_-$ for every $g$ and $\mathrm{BA}(g)=\tfrac12$. $\square$
The statement holds for any pair of laws, in particular for the empirical laws of $S$ on the positives
and negatives of a run: the observed balanced accuracy of every deterministic mark-only rule on those
items is at most $\tfrac12(1+\widehat{\mathrm{TV}})$, with $\widehat{\mathrm{TV}}$ the total variation
between the two empirical laws (for a randomised rule, its balanced accuracy averaged over its own
randomisation).

\section{Two Further Preregistered Tests}\label{app:further}

\paragraph{Explanation choice.} We scored all 105 Neuronpedia
explanations of the 49 features with Jev (2-level rubric) on both sets. The explanation that scores best on
stored snippets is also best on natural windows for 36 of 49 features (chance 0.48), and 90 of 105
explanations score higher on stored snippets; but two halves of one set disagree about the best explanation
(0.25, 0.31) about as often as the two sets do (0.29), so at ten windows per feature the scorer cannot
resolve the ranking, and a re-ordering is neither shown nor excluded.

\paragraph{A scorer trained on the cued data.} A small encoder trained on the stored snippets of 360 features
of the 500-feature sample to predict which tokens fire scores 0.526 on 100 held-out features (two seeds,
\cref{app:protocol}). The preregistered shortcut criterion, half the score with a blank explanation, is not
met (12\%); a blank explanation is out of distribution for it. The in-distribution controls say the rest: the
same model trained with no explanation scores 0.438 on the same snippets (83\% of the full model; both are
below the centre prior's 0.716 there, so the encoder learns the position only in part), a
shuffled explanation keeps 31\% of the full model's score (Jev: 6\%), and on natural windows the full model
falls to 0.193, by 63\% (Jev: 31\%).

\section{Protocol Details}\label{app:protocol}

\paragraph{Simulation questions.} The state of one request holds the explanation, a note (``Each
document is written as \texttt{<index>\guilsinglleft{}token\guilsinglright{}}; whitespace inside the brackets is part of the
token'') and the document rendered as \texttt{<0>\guilsinglleft{} edit\guilsinglright{}<1>\guilsinglleft{} ]\guilsinglright{}\dots}. Each token is one
\texttt{Score} question whose instructions are the object
\texttt{\{document, token\_index, token\}}, with one of the rubrics below; the read-out is the
returned expected level only, and the reported confidence is not used.
\begin{itemize}[leftmargin=*,itemsep=0pt,topsep=2pt]
\item 10 levels: ``0 - nothing to do with the description'', ``1 - a very faint, incidental
connection'', ``2 - a weak connection'', ``3 - a mild connection'', ``4 - a moderate connection'',
``5 - a clear connection'', ``6 - a clear and direct instance'', ``7 - a strong, central instance'',
``8 - a very strong, textbook instance'', ``9 - the strongest possible instance of the description''.
\item 4 levels: ``0 - not the described thing'', ``1 - weakly or indirectly related'', ``2 - clearly
the described thing'', ``3 - the strongest possible instance of the description''.
\item 2 levels: ``0 - not the described thing'', ``1 - the described thing''.
\end{itemize}
The API accepts at most 10 levels, so GPT-4's 0--10 becomes 0--9. By \cref{prop:affine}(a) the change
of numeric range does not matter; the change from eleven levels to ten is a change of rubric, which (a)
does not cover (the 2-, 4- and 10-level rubrics are compared in \cref{tab:sae}). The ``bare index'' arms render all tokens of the batch as one
\texttt{<n>\guilsinglleft{}token\guilsinglright{}} stream and use the integer $n$ as the question's only instruction. The ``terse''
arm uses ten bare numerals as criteria and moves the level wording into the state.

\paragraph{Detection and fuzzing questions.} The state holds the explanation and the excerpts as
plain text. Each excerpt gets one \texttt{Noul} (yes/no) question asking whether it contains at least one token
the feature would fire on, with the two criteria ``at least one token in this excerpt is an instance
of what \texttt{explanation} describes'' and ``no token in this excerpt is an instance of what
\texttt{explanation} describes''. Fuzzing wraps marked tokens in \texttt{<< >>} and asks whether the
marked tokens are instances of the explanation, thresholded at 0.5 before the run.

\paragraph{Batching and latency.} Production runs used batches of up to 70 questions per request and
eight concurrent workers. On the 400-neuron run (4{,}399 requests, 63.7 questions per request on average)
the measured median latency was 1.15\,s and the 95th percentile 2.93\,s per request, read from the run's
usage record. The vendor's documented limit of 1{,}200 requests per minute binds beyond about 23 workers.

\paragraph{Explanation-choice test.} All 105 Neuronpedia explanations of the 49 natural-window features (43
features carry two, 5 three, 1 four) were scored by Jev under one protocol on the paper's stored snippets and
on the natural windows, with the 2-level rubric as the preregistered primary and the 10-level rubric secondary.
Best-explanation agreement is the fraction of features whose top explanation coincides across the two sets;
the split-half disagreement draws 2{,}000 stratified half-splits of each set's windows and reports how often
the two halves disagree about the best explanation within a set, and between a half of each set. Re-scoring
the paper's explanation with the 10-level rubric on identical inputs reproduced 6.5\% (stored) and 9.8\%
(natural) of token-level values exactly, Pearson 0.997 and 0.995.

\paragraph{Trained scorer.} A 22.7M-parameter sentence encoder (all-MiniLM-L6-v2) with a linear token head
reads \texttt{[CLS] explanation [SEP] snippet [SEP]} and is trained with binary cross-entropy to predict whether
each GPT-2 token fires (activation above zero), on the stored snippets of 360 features of the 500-feature sample (40
more for epoch selection, 100 held out), for 6 epochs and two seeds (held-out scores differ by less than 0.01
between seeds). Each GPT-2 token's target and prediction attach to the first word piece overlapping its span.
The no-explanation arm is the same model trained with the explanation blank. Scores are the per-feature
pooled Pearson correlation of the sigmoid output with the raw activations, averaged over features.

\paragraph{Open-weight simulator.} Llama-3.1-8B-Instruct with the prompt of OpenAI's all-at-once simulator
(system message, three few-shot records and the target sequence as a prefilled assistant turn with every
activation \texttt{unknown}), rendered through the model's chat template; at each \texttt{unknown} slot the
prompt log-probabilities over the digit tokens 0--10 are renormalised and their expected value is the simulated
activation. One forward pass per sequence, nothing generated; 11{,}590 sequences (the 400 OpenAI neurons, the
49 natural windows and the 49 features' stored snippets, each with the feature's and a shuffled explanation).
The preregistered top-20 log-probabilities failed 21 of the 100 OpenAI sequences of the smoke run (a slot with
no digit among the top 20 fails the whole sequence), so the full run used the top 100, decided before it
started; 88 sequences still failed and were dropped, never zero-filled, and every unit kept at least nine of
its ten documents. Inference took 2\,h\,10\,min on one H200; the mean GPU utilisation was 4.6\%, since the
engine post-processes the top-100 prompt log-probabilities on the CPU. Reading the eleven digit logits at the
slot positions directly would avoid this.

\paragraph{Harm-test control.} After the real-data result, the preregistered rule of \cref{sec:harm} was run on
synthetic rankings of 49 features (post hoc): with stored scores built from the crop geometry alone plus noise it
returned ``harm demonstrated''; built from the natural-window scores alone, ``no ranking harm shown''; mixing the two,
it first fired at a geometry weight of 0.6 (4 of 5 seeds) and from 0.7 in every seed. This calibrates the instrument
on synthetic data and is not a positive control on real data.

\paragraph{Real-data control and crop-by-length factorial (preregistered).} On the natural corpus we rebuilt, for each of
the 49 natural-window features, 20 activating occurrences (one per document; fewer for 7 features; 6 features with under
10 usable occurrences dropped, leaving 43) and 5 non-activating windows, cropped four ways on the same occurrences:
peak-centred or at a random offset, 17 or 64 tokens. Jev's mean score is 0.462 (centred, 17), 0.471 (random, 17), 0.380
(centred, 64) and 0.385 (random, 64); the length effect is $+0.084$ [0.065, 0.103] (40 of 43 features), the crop effect
$-0.006$ [$-0.012$, $-0.001$] (30 of 43 lower), the interaction $-0.004$ [$-0.017$, 0.009]. So for Jev the level gap of
\cref{sec:harm} follows window length, not the centred crop. Stitched passages, 74 tokens long on average and carrying a
relocated centre cue, keep 0.476 (\cref{sec:clean}), a partial control for length only. The preregistered second scorer (Llama-3.1-8B-Instruct) was not
run, so the two-scorer rule is not decided for length; for the crop it cannot be met, since Jev's effect has the
opposite sign. For the positive control we inserted the centre token of each centred 64-token window a second time
(the copy carrying the same activation). This raises Jev's level by 0.063 [0.050, 0.076] (37 of 43) and its rank
correlation with the centre prior to 0.363 (random-offset windows: $-0.008$). Test scores took the cued value for a
random share $q$ of features and the random-offset value for the rest; the unchanged rule of \cref{sec:harm} returned
``harm demonstrated'' in 0 of 200 draws at each of $q=0.25$, 0.5 and 0.75, and also with every feature cued, where
$\rho_{\mathrm{geo}}=0.363$ and $\rho_{\mathrm{nat}}=0.895$. ``Geometry confound present'' fired in 8, 43 and 95 of 200
draws. By the preregistered rule the test is not sensitive at $q=0.5$. The cue raised levels but hardly moved the ranking,
so this control produced no reordering for the rule to find; the test's power against a reordering on real data remains
unmeasured.

\section{Cue Use and Cue Levels Across Dictionaries}\label{app:dicts}

\paragraph{Cue use by the stored scorers.} Preregistered before the analysis ran. On the 1{,}000 stored runs of
gemini-2.0-flash and llama-3.1-70b (500 each, on the same items; the one claude-3.5-sonnet run is left out), with
marks rebuilt as in \cref{sec:cues}, positives on which the centre (count) rule and the label disagree are
compared with positives of the same run on which they agree: Mantel--Haenszel risk difference stratified by run,
Cochran--Mantel--Haenszel test, run-cluster bootstrap interval (10{,}000 draws), Holm correction over the two
scorers and two cues. Outcomes: gemini-2.0-flash ``follows the cue'' for the centre cue ($-0.244$, Holm
$p=5.5\times10^{-21}$; 116 runs hold both kinds of positive) and accepts positives with the negatives' mark count
slightly more often ($+0.050$ [0.01, 0.09], Holm $p=0.021$); llama-3.1-70b shows neither ($-0.036$, Holm $p=0.55$;
$-0.005$ [$-0.05$, 0.04], Holm $p=0.87$). On the items where label and rule disagree, the verdicts match the label
on 0.64 and 0.70 (centre, 325 items per scorer) and 0.80 and 0.68 (count, 613 items, all positives: no negative
disagrees with the count rule), every Wilson interval above one half. Centre-marked negatives (30 per scorer,
from features whose negatives carry 14 to 17 marks) were accepted on 4 and 1 items, against 0.061 and 0.092 of the
other negatives (Fisher $p=0.11$ and 0.52); with the seed fixed, whether a negative is marked at the centre is set
by its mark count, so this comparison is confounded. Positives with the negatives' mark count are not accepted
less often than positives with one mark fewer ($+0.026$ and $-0.021$; $p=0.18$ and 0.42).

\paragraph{Per-feature correlations on the first 50 features.} On the stored fuzzing runs of the first 50 features, the preregistered Spearman correlation of each
scorer's (llama-3.1-70b, gemini-1.5-flash, gemini-2.0-flash) per-feature balanced accuracy with the centre
rule's is $-0.20$, $-0.17$ and $-0.19$ (bootstrap intervals include zero). Outside the preregistered test,
accuracy rises with the count rule's (Spearman 0.30, 0.31 and 0.31, bootstrap intervals excluding zero;
the Pearson correlations, 0.18, 0.11 and 0.19, have intervals that include zero) and, for the two Gemini scorers, with the simulation centre prior (0.42 and 0.50),
which could reflect use of a cue or how sharply a feature fires.

\paragraph{Cue levels across dictionaries.} Preregistered before any rule was computed on the new dictionaries.
Six Neuronpedia sources: the Gemma Scope layer-20 residual SAEs of Gemma-2-2B (16k and 65k latents) and Gemma-2-9B
(16k and 131k) \citep{lieberum2024gemmascope}, OpenAI's TopK SAE with 32k latents on GPT-2 small layer 8
\citep{gao2024scaling}, and the ReLU SAE of \cref{sec:cues} (the 500-feature sample). For the first five, 200
features each were drawn at random among those with at least 10 distinct hosted activating examples; of the pages
fetched for them, only the activation values and hashes of the token strings were kept. Positives are the hosted
examples cut to a window: the 32-token grid window that holds the peak (delphi's windows, peak anywhere; primary),
the 17-token window centred on the peak (Neuronpedia's GPT-2 crop), or the context as hosted. Marks follow the
fork's rule (\cref{sec:cues}); negatives are all-zero windows marked with the fork's seeded sample. The primary
comparison, fixed in advance, is the count rule on 32-token windows for Gemma Scope 9B, 131k minus 16k, the pair
with a published fuzzing gap (0.76 against 0.73, medians; \citealp[Table A10]{paulo2024autointerp}): ``as large as
the published gap'' if the difference is at least 0.03 in size and its bootstrap interval excludes zero, ``equal to
within it'' if the interval lies inside $\pm0.03$, otherwise ``not resolved''. Outcome: $-0.028$ [$-0.041$, $-0.015$],
not resolved. The sign is opposite to the published gap because the wider dictionary's positives carry fewer
active tokens (the negatives' mark count is 4.1 against 7.1). On the same pair and windows the centre rule, a
secondary comparison, differs by $-0.015$ [$-0.027$, $-0.002$], equal to within the published gap. So on this pair the cue
that differs is one neither stored scorer follows, and the cue one of them follows does not differ. Across the five
dictionaries the count rule's level spans 0.925 to 0.969, more than published fuzzing gaps between SAEs. Validation: on the 500 res-jb features the rebuilt
17-token items give count and centre rule scores of 0.964 and 0.989, against 0.969 and 0.979 on the stored runs;
on the 12 Gemma-2-2B features with stored runs the hosted contexts give 0.968 and 0.528, against 0.981 and 0.552.
On 17-token crops the two GPT-2 small layer-8 SAEs are equal on the count rule ($+0.002$, TopK minus ReLU) and differ
by $-0.106$ on the centre rule (\cref{tab:dicts}). These are levels of explanation-free rules on rebuilt items: the
size of a cue difference a scorer could pick up, not a difference in any scorer's score.

\paragraph{Matched-mark rescoring.} Preregistered before any item was built or scored. The fork's prompt (system
message, three few-shot exchanges, five items per call), parser and sampling (temperature 1), with open weights run
locally: Llama-3.1-70B-Instruct, the stored llama scorer's model, and gemma-3-27b-it in place of gemini-2.0-flash,
which its provider shut down in June 2026. Each item is scored twice, with the fork's marks and with matched marks;
text, order, batches, explanation and sampling seed are the same, and only the negatives' marks differ. On the
500-feature runs each negative takes the marks of a positive of the same length. On Gemma Scope 9B, 190 (16k) and 198
(131k) of the 200 features above have 10 activating examples, cut to 32-token grid windows, and 10 windows of hosted
contexts on which the feature is zero throughout; the explanation is the page's gpt-4o-mini one, and the positives'
mark sets are reassigned one-to-one, so every mark-only rule scores exactly $\tfrac12$. Primary outcome: the change in
the 131k minus 16k gap, ``changes the ordering'' if it is at least 0.03 in size with an interval excluding zero (or the
gap changes sign), ``does not change it'' if its interval lies inside $\pm0.03$. Llama: 16 of 8{,}102 calls fail to
parse; it scores 0.861 on the original items against 0.812 for the stored runs, a difference of 0.049 against the
preregistered reproduction margin of 0.05 (a point estimate just inside it, not an equivalence test; one sample, a
different runtime); its level changes by $-0.001$ [$-0.005$, 0.003] on the 500 features, $+0.004$ on 16k and $+0.008$ [0.001, 0.015] on
131k, and its true-negative rate on the 500 features is 0.973 against 0.974; the gap is $-0.025$ [$-0.056$, 0.008] with the fork's
marks and $-0.021$ [$-0.052$, 0.011] with matched ones, a change of $+0.004$ [$-0.007$, 0.015]: inside $\pm0.03$, so by the
preregistered rule the cue does not change this gap. Both gap intervals include zero and both point estimates are
opposite in sign to the published $+0.03$, so this local scorer does not reproduce the published ordering, and the
result says only that removing the mark cues leaves its gap where it was. gemma-3-27b-it:
2{,}017 calls (25\%) fail the fork's parser: 1{,}385 hold malformed lists (for example stray text or doubled commas), 331 no closed
list, 210 a trailing comma and 91 a list of the wrong length, and 96 further items carry a value other than 0 or 1.
A feature enters a comparison only if both constructions give it a balanced accuracy, so gemma-3-27b-it's comparisons
rest on 386 of the 500 features, 151 of 190 (16k) and 156 of 198 (131k). Its level rises by 0.028 [0.018, 0.038]
on the GPT-2 features, 0.035 on 16k and 0.018 on 131k, through the true-negative rate (0.870 to 0.923); its gap change is
$-0.017$ [$-0.039$, 0.004], not resolved. With the fork's marks both accept positives whose centre is unmarked less
often than centre-marked ones of the same feature ($-0.088$ and $-0.162$; $p=0.0019$ and $2.1\times10^{-11}$). Per-feature
scores under the two constructions have Spearman correlation 0.95 (llama) and 0.87.

\begin{table}[!ht]
\centering\footnotesize\setlength{\tabcolsep}{3pt}
\caption{Balanced accuracy of the count and centre fuzzing rules, mean over features with 95\% bootstrap intervals,
on items rebuilt from Neuronpedia's hosted examples with the scorer fork's marks (Gemma Scope: layer 20; GPT-2 small:
layer 8). C32: the 32-token grid window
holding the peak; C17: the 17-token window centred on the peak (the res-jb SAE is hosted as 17-token crops, so it
has no C32 row).}
\label{tab:dicts}
\begin{tabular}{lrcccc}
\toprule
SAE & features & C32 count & C32 centre & C17 count & C17 centre \\
\midrule
Gemma-2-2B, Gemma Scope 16k & 200 & 0.957 [0.946, 0.965] & 0.542 [0.530, 0.552] & 0.953 [0.942, 0.961] & 0.934 [0.923, 0.943] \\
Gemma-2-2B, Gemma Scope 65k & 200 & 0.944 [0.934, 0.952] & 0.543 [0.532, 0.554] & 0.932 [0.922, 0.942] & 0.929 [0.917, 0.939] \\
Gemma-2-9B, Gemma Scope 16k & 200 & 0.953 [0.945, 0.960] & 0.560 [0.548, 0.570] & 0.940 [0.932, 0.948] & 0.946 [0.938, 0.953] \\
Gemma-2-9B, Gemma Scope 131k & 200 & 0.925 [0.913, 0.936] & 0.544 [0.538, 0.551] & 0.914 [0.901, 0.927] & 0.948 [0.939, 0.955] \\
GPT-2 small, OpenAI TopK 32k & 200 & 0.969 [0.962, 0.975] & 0.524 [0.516, 0.533] & 0.966 [0.959, 0.973] & 0.882 [0.875, 0.889] \\
GPT-2 small, res-jb (ReLU) & 500 & -- & -- & 0.964 [0.959, 0.968] & 0.989 [0.985, 0.992] \\
\bottomrule
\end{tabular}

\end{table}

\section{Data Sources}\label{app:provenance}

Neuronpedia's stored runs (GPT-2 small layer 8: 500 random features, plus the first 50 features in index
order, each with the explanation and the 25 snippets of its stored fuzzing runs, where runs cover more than one
explanation the one with the most runs; Gemma-2-2B layer 20: 12 features) and the activations of the stitched
passages were obtained from its public API in September 2026 \citep{neuronpedia}. The delphi items were
built with our re-implementation of the example constructor, sampler and fuzzing marks of EleutherAI/delphi at commit
\texttt{4fea06e} \citep{eleuther2024delphi} (transcribed from its source, with its data classes replaced by plain arrays;
delphi was not run as a package) and SAEBench's scoring
windows with SAEBench at commit \texttt{8042bb3} \citep{karvonen2025saebench}, both on the 10k-document
OpenWebText subset (OpenWebText-10k). The 400 GPT-2 XL neurons come from OpenAI's public release
\citep{bills2023neurons}. The natural windows and the BOS recomputations use GPT-2 small and the same SAE
\citep{bloom2024gpt2saes}, run locally with the BOS token. Which fork commit Neuronpedia deployed for the stored runs cannot be read: its 2025 build installed the fork by URL without a
revision, and its current lock pins \texttt{05d5718}, whose scorer code is identical to the fork's head from February 2025 on.
We installed that commit with Neuronpedia's locked dependencies in a separate environment and ran Neuronpedia's fuzzing route
on every stored run we use (2{,}030 runs), with the LLM call replaced by a stub that records the prompt (preregistered). Its
marks equal our re-implementation's on all 50{,}699 items, as do the example texts, the negatives' mark count on every run and
all 10{,}146 prompts. The fork shuffles the items with the random state left by its last seed reset, so the stored item order
is a function of the run's composition and mark count; the fork reproduces it on all 2{,}030 runs (with the mark count off by
one, on 18\% of runs). The fork's head before February 2025 gives the same result on the 500-feature sample. The order test
checks the seed and the negatives' mark count against the stored data; the positives' 0.3 threshold uses no randomness and is
checked against the fork's code only. For the delphi items of \cref{sec:cues}: the branch of \texttt{center\_examples} is inverted relative to its name
since delphi commit \texttt{cc9387c}; with the flag
set to \texttt{False}, which centres the windows, the peak sits at index 24, the fixed-index simulation prior scores 0.466, the rule
at index 24 is cancelled (0.491), and the companion mark at index 31 still gives 0.994 (chosen after the
fact); the count rule gives 0.935. Stored accuracies are recomputed from the per-item verdicts; the score Neuronpedia displays differs
from this recomputation on 112 of the 1{,}001 runs (its mean is 0.845 against 0.838), and the centre rule is
above the displayed score on 735 runs and below it on 89. Each stored item also has a \texttt{correct} field. On 172 of the
25{,}011 items it disagrees with the item's verdict compared against its label. We score the verdict, which is the
scorer's own output, and not \texttt{correct}, which should be derived from it; Neuronpedia's formula applied to
\texttt{correct} gives 0.821.
For the cross-dictionary levels (\cref{app:dicts}),
feature pages of four further sources were fetched from the same API in October 2026. The 1{,}001 GPT-2 runs by LLM scorers split
(500 by gemini-2.0-flash, 2025-02, 500 by llama-3.1-70b-instruct, 2025-03, and 1 by claude-3.5-sonnet), and the Gemma
features carry 27 newer runs (2025-08 to 2026-03). The cases of \cref{fig:cases} follow rules fixed before their outcomes were seen. Marks: the first
gemini-2.0-flash run in sample order with that scorer's median stored accuracy, the most common count-rule accuracy and
the most common negative mark count. Position: the first run holding both a centre-marked positive and a positive with
the centre unmarked. Sequence order: the neuron whose file-order rule minus GPT-4 is closest to the median over the 400.
Clean window: the feature with the median Jev $r$, and its window with the largest activation. Not preregistered$^\dagger$: delphi draws the explanation and test examples with the
same reset seed, so 10 of the 50 test examples of every feature were also shown to the explainer.

\paragraph{Count rules without the negatives' mark count.} Chosen after the result of the count rule was seen$^\dagger$.
The fork's $n$ is the mean number of active tokens on a run's positives, rounded up, and the count rule of
\cref{sec:cues:marks} is told it. On the same runs and rebuilt marks we scored three rules that are not:
(a) ``yes'' if $|M(x)|$ is below the largest mark count in the run, which equals $n$ on 875 of the 1{,}001 GPT-2 runs and
19 of the 27 Gemma runs; (b) ``yes'' if $|M(x)|$ differs from the run's most common mark count; (c) ``yes'' if $M(x)$ is
one contiguous span. Rules (a) and (b) read the unlabeled marks of the whole run, rule (c) those of the item; none reads a
label, an activation or the explanation. On GPT-2 they reach 0.905, 0.173 and 0.924 against the stored 0.838
(differences $+0.067$ [0.050, 0.084], $-0.665$ [$-0.694$, $-0.636$] and $+0.086$ [0.072, 0.099]; rules (a) and (c) are above
the stored score on 653 and 648 runs and below it on 208 and 194). On Gemma they reach 0.837, 0.442 and 0.730 against
0.794 ($+0.044$ [$-0.134$, 0.171], $-0.352$ [$-0.617$, $-0.089$] and $-0.064$ [$-0.154$, 0.045]). Means are over runs, as
for the count rule; intervals are 95\% bootstrap intervals over features (10{,}000 resamples). None of the three is above the stored scorers on both populations. Rules (a) and (c) were then frozen and tested, as
preregistered, on 500 held-out features of the same dictionary (1{,}001 runs by the same scorer batches; items hashed before
scoring): they reach 0.902 and 0.913 against the stored 0.844 ($+0.058$ [0.040, 0.077] and $+0.069$ [0.055, 0.084]; above the
stored score on 640 and 625 runs, below on 236 and 236), within 0.03 of their first values, and above each scorer separately
(gemini-2.0-flash $+0.031$ [0.011, 0.050] and $+0.043$ [0.027, 0.059]; llama-3.1-70b $+0.085$ [0.066, 0.104] and $+0.097$
[0.082, 0.112]); the count rule told $n$ reaches 0.959. A probe of 21 other dictionaries and 3{,}000 further Gemma Scope
features found one held-out run, on Gemma-2-2B layer 0, too few to test; the Gemma result above stays post hoc.

\section{Preregistrations and Their Outcomes}\label{app:prereg}

\paragraph{Extension to 400 neurons.} Written after the 40-neuron pilot and before the extension ran:
per-neuron pooled $r$, Jev 10-level minus GPT-4 published, sign test, with three outcomes worded in
advance; no exclusion, no rubric choice after the fact, no change of pooling. Outcome: significantly
higher.

\paragraph{Follow-up analyses.} Two sets of follow-ups were preregistered with decision rules before
they ran, and their outcomes appended without editing the rules (in the preregistration file, to be released with the
code). The first: rank correlation, sequence-level terms, pilot exclusion, Holm correction and equivalence
intervals; the fuzzing position and count rules; BOS versus context in the local recompute; natural
windows; the exact piece-centre prior; the causal-prefix and 2-level-rubric checks; and a current LLM
simulator, which was not run. The second: the harm test of \cref{sec:harm}; the delphi and SAEBench
builds; the 500-feature analysis on Neuronpedia; and small recomputes (equivalence on rebuilt marks, the
natural-window split, the chunking of the natural windows, the first BOS-less attempt, the BOS
mechanism). The decision rules and verdicts: harm, ``geometry confound present, ranking harm not
shown''; 500-feature sample, ``leak replicates'' on GPT-2 (the same rebuild rule on 500 further features) and ``no position leak'' on Gemma (the preregistrations' ``leak'' is this paper's cue), with the count rule
reported regardless; delphi default, ``no fixed-offset leak in simulation'' and a fuzzing position cue
above the 0.6 bound; BOS mechanism, ``stated as measured''; rank correlation, ``does not separate them''
for the 10-level rubric and, for the 4-level rubric, ``GPT-4 is higher on rank correlation'' (Spearman
$-0.016$, $p=0.014$; \cref{tab:h2h}). The first BOS-less attempt used a subsample of 8 features and 8
snippets (0.138, against 0.082 on all features). Without the BOS token, dropping the first position
restores a median correlation of 0.985, and centring each position's residual restores 0.994 with the
norm ratio unchanged.

\paragraph{Third set.} Written before each item ran, outcomes appended: the explanation-choice test (rule:
best-explanation disagreement of at least 0.25 between the two sets with the split-half test at $p<0.05$;
outcome ``not shown'', with the instrument flag that the scorer cannot resolve the ranking at this sample
size; the 10-level rubric, run as the secondary arm, gives 37 of 49 features and 93 of 105 explanations); the
trained scorer (rule: at least half the score kept with a blank explanation and at least half lost
on natural windows, on the two-seed mean; outcome ``not shown'', the first clause failing at 12\%); and the
open-weight simulator (rule: above its shuffled control on the natural windows with a sign test at $p<0.05$;
outcome ``reads the explanation'', instrument sensitive on the OpenAI neurons; lower than Jev on all three
sets; against GPT-4 not resolved in pooled correlation and lower in rank correlation). Its one deviation, the
top-100 log-probabilities, was decided on the smoke run before the full run (\cref{app:protocol}).

\paragraph{Fourth set.} Written before any code for them ran, outcomes appended: cue use by the stored scorers
(outcomes per scorer and cue) and the cue's level across dictionaries (primary outcome ``not resolved''), both in
\cref{app:dicts}. The explainer comparison in \cref{sec:harm} is the third preregistered metric of the
explanation-choice test.

\paragraph{Not preregistered.} The train/test overlap in delphi, found while replicating its sampler; the
paired level comparison and the count-rule and simulation-centre-prior correlations of the stored fuzzing
accuracy (\cref{sec:harm}); the offset shared by all features on the natural windows, computed after the preregistered
natural-window analyses (\cref{sec:clean}); the additive within/between split of \cref{sec:h2h}; the
centre prior and $\eta$ on the 50 stored-snippet features and the 400 OpenAI neurons, computed before the follow-up preregistrations; and two
analyses of existing outputs, each recorded as post hoc in the preregistration file: the sequence-order rule
on OpenAI's release and the centre rule with uniformly random negative mark positions (\cref{sec:cues}).

\paragraph{Deviations for the natural windows.} All three were fixed before any Jev call: the corpus
was read from the loader's source archive; documents were truncated to 1{,}016 tokens and cut into
127-token chunks, each run with the BOS token as a 128-token context (the SAE's training length), so every
window has at most 127 tokens of left context; and a first build that drew each offset from the
occurrence's feasible range gave a U-shaped peak histogram and was discarded before scoring.

\paragraph{Open typed decision models.} Preregistered before running: both 1{,}024-token Laya
checkpoints, \texttt{laya-typed-decisions} (421M) and \texttt{laya-multilingual} (322M), on the 400 neurons with
byte-identical inputs. Both fell below GPT-4 and were
indistinguishable from their own shuffled controls (\cref{tab:laya}); a rendering change tested with a
written rule did not repair them. We draw no conclusion about typed decision models as a category.

\begin{table}[!ht]
\centering\small
\caption{Two open typed decision models on the simulation protocol. Same 400 neurons, inputs
byte-identical to Jev's; the shuffled-explanation control uses a seeded 100-neuron subset.}
\label{tab:laya}
\begin{tabular}{lccc}
\toprule
scorer & mean $r$ & $\Delta$ vs GPT-4, sign test & $\Delta$ vs shuffled, sign test \\
\midrule
GPT-4, published & +0.160 & -- & -- \\
Jev & +0.197 & +0.037, $p=2.3\times10^{-8}$ & -- \\
\texttt{laya-multilingual} (322M) & +0.012 & $-$0.148, $p=8.6\times10^{-25}$ & +0.005, $p=0.37$ \\
\texttt{laya-typed-decisions} (421M) & +0.000 & $-$0.160, $p=9.0\times10^{-26}$ & +0.012, $p=0.057$ \\
\bottomrule
\end{tabular}

\end{table}

\section{Stitched Passages}\label{app:stitch}

Each passage concatenates four snippets in a seeded random order, one to three from the feature
(cycling with the passage index) and the rest from other features, and its activations come from
Neuronpedia's activation endpoint on the stitched text: 49 features, 392 passages. Each feature gets
eight passages; one of the 50 features has fewer than three activating snippets and was skipped. The
peaks sit at the four piece centres, so a prior at the true centres scores 0.496, and Jev is not above it
(\cref{sec:clean}). The same scorer's 0.476 on stitched
passages overstated its natural-window score by 0.146 (lower on natural windows for 43 of 49 features,
$p=5.7\times10^{-8}$). Snippet tokens were
joined as text and re-tokenised by the endpoint, and 359 of the 392 passages contain byte-level
mojibake. The rubric ablation below was run on this set; the comparison of the 10- and 4-level rubrics
does not survive Holm correction.

\begin{table}[!ht]
\centering\footnotesize\setlength{\tabcolsep}{3pt}
\caption{Rubric and encoding ablation on the 49 stitched features, measured input tokens per question,
paired per feature against the 4-level rubric, with Holm-adjusted $p$. ``Bare index'' drops the token
text from the question. The true piece-centre prior (0.496) is within 0.01 of the best arm, so the arms are
compared on items that carry a relocated position cue (\cref{sec:clean}).}
\label{tab:sae}
\begin{tabular}{lcccccc}
\toprule
Jev condition & tokens/q & mean $r$ & $\Delta$ vs 4 levels & W / L & $p$ & Holm $p$ \\
\midrule
2 levels & 69.3 & 0.502 & +0.005 & 28 / 21 & $0.39$ & $0.39$ \\
4 levels & 97.3 & 0.497 & -- & -- & -- & -- \\
10 levels & 165.3 & 0.476 & $-$0.020 & 17 / 32 & $0.044$ & $0.089$ \\
10 numerals, legend in state & 110.4 & 0.463 & $-$0.034 & 13 / 36 & $0.0014$ & $0.0042$ \\
2 levels, bare index & 47.2 & 0.341 & $-$0.155 & 7 / 42 & $3.6\times10^{-7}$ & $1.8\times10^{-6}$ \\
4 levels, bare index & 75.2 & 0.335 & $-$0.161 & 8 / 41 & $2.0\times10^{-6}$ & $7.9\times10^{-6}$ \\
\midrule
piece-centre prior, true boundaries & -- & 0.496 & -- & -- & -- & -- \\
centre prior & -- & $-$0.021 & -- & -- & -- & -- \\
shuffled explanation & -- & 0.021 & -- & -- & -- & -- \\
\bottomrule
\end{tabular}

\end{table}

\section{Detection and Fuzzing on the First 50 Features}\label{app:det}

\begin{table}[!ht]
\centering\small
\caption{Detection and fuzzing on the 50 stored-snippet features. Stored scorers: Neuronpedia's stored
fuzzing runs, balanced accuracy averaged over runs; the rules and Jev's fuzzing use marks rebuilt from
the scorer fork's code. Detection shows items unmarked, so no mark rule applies; Neuronpedia stores no
detection runs for these scorers (two runs of a different recall-type score for gemini-1.5-flash were not
used), so Jev's detection has no stored counterpart. Jev's threshold is 0.5,
set before the run; its rows are means over the 50 features of per-feature balanced accuracy and AUROC.}
\label{tab:det}
\begin{tabular}{lcc}
\toprule
scorer or rule (items) & balanced accuracy & AUROC \\
\midrule
Jev detection (unmarked) & 0.803 & 0.893 \\
Jev detection, shuffled explanation & 0.504 & 0.499 \\
Jev fuzzing (rebuilt marks) & 0.797 & 0.949 \\
gemini-1.5-flash, stored fuzzing (54 runs) & 0.858 & -- \\
gemini-2.0-flash, stored fuzzing (51 runs) & 0.841 & -- \\
llama-3.1-70b-instruct, stored fuzzing (50 runs) & 0.790 & -- \\
centre rule (rebuilt marks) & 0.985 & -- \\
count rule (rebuilt marks) & 0.956 & -- \\
\bottomrule
\end{tabular}

\end{table}

On the first 50
features (\cref{tab:det}) the same rule reaches 0.985 against 0.790, 0.858 and 0.841 for llama-3.1-70b, gemini-1.5-flash
and gemini-2.0-flash (sign tests $p\le4.1\times10^{-5}$), and the centre prior 0.723, which
\cref{prop:centre} reproduces per feature to $10^{-14}$ (an algebraic identity on the same data); the centre token carries on average
68\% of each feature's activation mass there.

On rebuilt marks Jev's fuzzing accuracy is equivalent to llama-3.1-70b's within 0.05 (difference
$+0.0065$, 90\% interval $[-0.035, 0.048]$), and both are below the centre and count rules. These
comparisons are on items that carry the cues of \cref{sec:cues}, so they order scorers on the same data
but do not measure explanation quality.

\section{Cost}\label{app:cost}

Jev bills input tokens only, but a per-token protocol re-sends the rubric with every token: 69.3 input
tokens per token scored with the 2-level rubric and 165.3 with the 10-level one. The literature's
costs for Llama-70b scorers \citep{paulo2024autointerp} are estimates for 100 contexts per feature. At
that budget with 32-token contexts, Jev simulation costs \$931 per 100k features with the 2-level rubric
against an estimated \$3{,}600 for all-at-once simulation, a factor of 3.9, and \$2{,}222 with the
10-level rubric, a factor of 1.6; detection costs \$58 against an estimated \$588, a factor of 10.1. At
200 contexts the 10-level rubric costs \$4{,}443 per 100k features, 1.23 times the 100-context estimate,
that is, more than the LLM; since that estimate is for half the budget, the same-budget factors above are
the comparison. The 2-level rubric is weaker on neurons: above GPT-4 ($+0.0187$, $p=5.4\times10^{-4}$) but
below the 10-level rubric ($-0.0186$, $p=7.1\times10^{-9}$; \cref{tab:h2h}). \Cref{tab:cost} projects whole
dictionaries at the 640-token budget of \citet{bills2023neurons} with the 2-level rubric.

\begin{table}[!ht]
\centering\small
\caption{Projected cost and wall-clock time for whole dictionaries at the 640-token budget of
\citet{bills2023neurons}, 2-level rubric, at \$0.042 per million input tokens (output is not billed).
Wall clock is the larger of requests $\times$ median latency (1.15\,s, measured on the 400-neuron run,
\cref{app:protocol}) / 8 workers and the documented rate limit. The 425{,}984-feature row is a 16k-feature SAE on each of the 26 layers of Gemma-2-2B
\citep{lieberum2024gemmascope}.}
\label{tab:cost}
\begin{tabular}{lrrrr}
\toprule
dictionary (features) & simulation \$ & simulation h & detection \$ & detection h \\
\midrule
16{,}384 & 30.52 & 6.0 & 1.22 & 0.7 \\
24{,}576 & 45.78 & 9.0 & 1.82 & 1.0 \\
65{,}536 & 122.08 & 24.0 & 4.87 & 2.6 \\
100{,}000 & 186.28 & 36.6 & 7.43 & 4.0 \\
425{,}984 & 793.52 & 155.9 & 31.63 & 17.1 \\
\bottomrule
\end{tabular}

\end{table}

\newpage
\section{Head-to-Head Details}\label{app:h2h}

\begin{figure}[!ht]
\centering
\includegraphics[width=\textwidth]{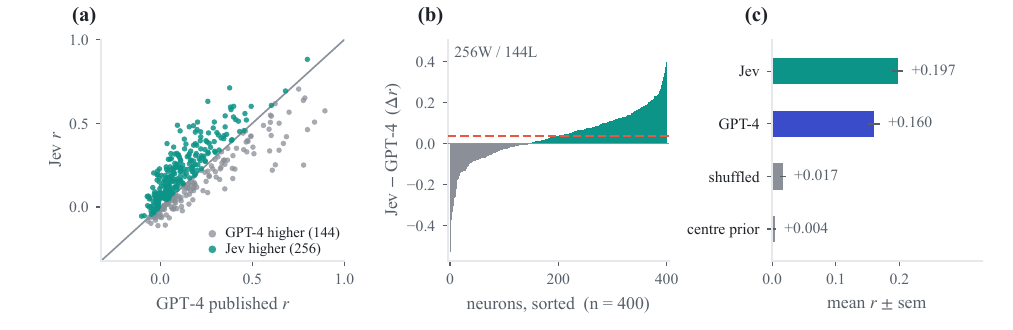}
\caption{Per-neuron pooled $r$ of Jev (10-level rubric) against GPT-4's published simulator on the 400
GPT-2 XL neurons. (a)~Per neuron; the line is parity. (b)~Paired difference, sorted; the dashed line is
the mean. (c)~Means $\pm$ sem beside the shuffled-explanation and centre-of-sequence controls.}
\label{fig:h2h}
\end{figure}

\paragraph{Pilot and pooling convention.} Recomputing GPT-4's pooled correlation from its stored
predictions reproduces the published score on every neuron to within $2\times10^{-15}$ (a check of the
pooling convention, not of the simulator). The primary test and the wording of its outcomes were fixed after a
40-neuron pilot ($\Delta=+0.025$, $p=0.27$) and before the run. The preregistered ``significantly higher''
outcome holds without the 40 pilot neurons (232 of 360,
$p=4.6\times10^{-8}$; \cref{tab:h2h,fig:h2h}).

\paragraph{Between and within sequences.} Jev spreads sequence means more ($\hat\eta$ 0.393
against 0.296); in the post hoc additive split of \cref{prop:affine}(b) the within-sequence term carries $+0.010$ of the
$+0.037$ gap (235 of 400 neurons, $p=5.4\times10^{-4}$) and the between-sequence term $+0.027$ (260 of
400, $p=2.0\times10^{-9}$). Even $r_w$ alone is higher for Jev (0.165
against 0.143, 242 of 400, $p=3.1\times10^{-5}$).

\paragraph{Causal prefix.} Jev sees the whole sequence, GPT-4 the prefix; restricting Jev to the causal prefix on
100 neurons changes its score from 0.185 to 0.182 ($-0.0023\pm0.0023$; median correlation between the two
predictions 0.95), an order of magnitude below the 0.037 gap, although on this subset neither variant is
significantly above GPT-4 (causal $p=0.13$, full context $p=0.76$).

\paragraph{Open 8B simulator.} The 8B open simulator scores 0.125 on these neurons: not resolved against GPT-4 at $p=0.05$ in
pooled correlation (182 of 400, $p=0.08$; lower in rank correlation) and below Jev (102 of 400).

\paragraph{Open 8B models through Jev's protocol.} Preregistered before any GPU run: Llama-3.1-8B-Instruct and Qwen3-8B
\citep{yang2025qwen3} answer, for each token, the typed 10-level \texttt{Score} question Jev answers, through AnyJev 0.2.0
\citep{zhang2026anyjev}; the read-out is the expected level after AnyJev's batch-prior correction, on the 400 neurons
with GPT-4's explanations and with shuffled ones. Both read the explanation (0.068 and 0.072, against 0.014 and 0.030
shuffled; higher on 295 and 285 of 400 neurons) and score below GPT-4 (by 0.092 [0.077, 0.108] and 0.087 [0.071, 0.105])
and below Jev (by 0.129 and 0.125), so the protocol alone does not explain Jev's lead. On the same Llama weights and the 4{,}038 of 4{,}082 sequences both runs completed,
the typed per-token protocol scores 0.057 [0.048, 0.067] below the all-at-once read-out above (282 of 400 neurons lower).
Deviations, logged before any full result was read: a fix to a doubled BOS token in the inference wrapper; 32-bit
instead of 16-bit floating point, because a preregistered agreement check between backends was tighter than 16-bit
precision; a transformers backend with a document-shared prefix after the 32-bit vLLM run failed; and about 21 H200 GPU-hours against an estimate
of 1 to 6. A preregistered optional 32B model was not run.

\begin{table}[!ht]
\centering\small
\caption{\textbf{Jev leads GPT-4's published simulator by 0.037 in pooled $r$, mostly between sequences,
and the lead is not detected in rank correlation.} 400 GPT-2 XL MLP neurons sampled before any score was
read. The top two blocks and the robustness block were preregistered (the 4-level Spearman row is the
preregistered adverse branch); the last block is post hoc: the within-sequence correlation $r_w$, the
$n_s$-weighted correlation of sequence means $r_b$ and the two additive terms of \cref{prop:affine}(b).}
\label{tab:h2h}
\begin{tabular}{lcc}
\toprule
scorer & mean $r$ $\pm$ sem & median $r$ \\
\midrule
GPT-4, published simulator & +0.160 $\pm$ 0.010 & +0.099 \\
Jev, 10-level rubric & +0.197 $\pm$ 0.009 & +0.169 \\
Jev, 4-level rubric & +0.191 $\pm$ 0.009 & +0.161 \\
Jev, shuffled explanation & +0.017 $\pm$ 0.004 & +0.007 \\
centre-of-sequence prior & +0.004 $\pm$ 0.002 & $-$0.009 \\
\midrule
paired, per neuron & $\Delta$ $\pm$ sem & W / L, sign test \\
\midrule
Jev (10-level) $-$ GPT-4 & +0.0373 $\pm$ 0.0061 & 256 / 144, $p=2.3\times10^{-8}$ \\
Jev (4-level) $-$ GPT-4 & +0.0313 $\pm$ 0.0057 & 251 / 149, $p=3.8\times10^{-7}$ \\
Jev $-$ shuffled explanation & +0.1800 $\pm$ 0.0099 & 334 / 66, $p=3.3\times10^{-44}$ \\
\midrule
\multicolumn{3}{l}{\emph{robustness (preregistered follow-ups)}} \\
\midrule
Jev (10-level) $-$ GPT-4, pilot excluded ($n=360$) & +0.0387 $\pm$ 0.0063 & 232 / 128, $p=4.6\times10^{-8}$ \\
Jev (2-level) $-$ GPT-4 & +0.0187 $\pm$ 0.0058 & 235 / 165, $p=5.4\times10^{-4}$ \\
Jev (2-level) $-$ Jev (10-level) & $-$0.0186 $\pm$ 0.0024 & 142 / 258, $p=7.1\times10^{-9}$ \\
Jev causal prefix $-$ GPT-4 ($n=100$) & +0.0181 $\pm$ 0.0117 & 58 / 42, $p=0.13$ \\
Jev causal $-$ Jev full context ($n=100$) & $-$0.0023 $\pm$ 0.0023 & 51 / 49, $p=0.92$ \\
Spearman: Jev (10-level) $-$ GPT-4 & $-$0.0092 $\pm$ 0.0041 & 180 / 220, $p=0.051$ \\
Spearman: Jev (4-level) $-$ GPT-4 & $-$0.0160 $\pm$ 0.0043 & 175 / 225, $p=0.014$ \\
\midrule
\multicolumn{3}{l}{\emph{post hoc, \cref{prop:affine}(b): within-sequence $r_w$; $n_s$-weighted correlation of sequence means $r_b$}} \\
\midrule
$r_w$: Jev 0.165 vs GPT-4 0.143 & +0.0217 $\pm$ 0.0054 & 242 / 158, $p=3.1\times10^{-5}$ \\
$r_b$: Jev 0.322 vs GPT-4 0.270 & +0.0522 $\pm$ 0.0161 & 229 / 171, $p=0.0043$ \\
within-sequence term $r_w\sqrt{1-\eta^2}\sqrt{1-\hat\eta^2}$ & +0.0099 $\pm$ 0.0050 & 235 / 165, $p=5.4\times10^{-4}$ \\
between-sequence term $r_b\eta\hat\eta$ & +0.0275 $\pm$ 0.0030 & 260 / 140, $p=2.0\times10^{-9}$ \\
\bottomrule
\end{tabular}

\end{table}

\end{document}